\documentclass[10pt,journal,compsoc]{IEEEtran}
\ifCLASSOPTIONcompsoc
  \usepackage[nocompress]{cite}
\else
  \usepackage{cite}
\fi
\ifCLASSINFOpdf
\else
\fi
\usepackage{amsmath,amsfonts,bm}

\def\eqref#1{equation~\ref{#1}}

\def\1{\bm{1}}

\def\vs{{\bm{s}}}

\DeclareMathAlphabet{\mathsfit}{\encodingdefault}{\sfdefault}{m}{sl}
\SetMathAlphabet{\mathsfit}{bold}{\encodingdefault}{\sfdefault}{bx}{n}

\usepackage{tikz}
\usepackage{pgfplots}
\pgfplotsset{compat=default}
\usetikzlibrary{pgfplots.groupplots}
\usepackage{graphicx}
\usepackage{amsmath}
\usepackage{amssymb}
\usepackage{booktabs}
\usepackage{algpseudocode}
\usepackage{algorithm}
\usepackage{subcaption}
\usepackage{bbm}
\usepackage{pifont}
\usepackage{multirow}
\usepackage{tabu}
\usepackage{multicol}
\usepackage{booktabs}

\newcommand{\xmark}{\ding{55}}
\newcommand{\cmark}{\ding{51}}%

\usepackage[pagebackref,breaklinks,colorlinks]{hyperref}

\usepackage[capitalize]{cleveref}
\crefname{section}{Sec.}{Secs.}
\Crefname{section}{Section}{Sections}
\Crefname{table}{Table}{Tables}
\crefname{table}{Tab.}{Tabs.}

\usepackage{xspace}

\makeatletter
\DeclareRobustCommand\onedot{\futurelet\@let@token\@onedot}
\def\@onedot{\ifx\@let@token.\else.\null\fi\xspace}

\def\eg{\emph{e.g}\onedot} 
\def\ie{\emph{i.e}\onedot} 
 
 \def\vs{\emph{vs}\onedot}

\makeatother

\makeatletter
\@namedef{ver@everyshi.sty}{}
\makeatother

\begin{document}
%
\title{Quantifying and Learning Static vs. Dynamic Information in Deep Spatiotemporal Networks}
%
%
%
%

\author{
{\small $^{1}$York University, $^{2}$Vector Institute for AI, $^{3}$Ryerson University, $^{4}$Samsung AI Centre Toronto, $^{5}$University of Guelph}
\\
{\small 
\texttt{\{m2kowal,msiam,wildes,kosta\}@eecs.yorku.ca}, \texttt{mdamirul@ryerson.ca}, \texttt{brucen@uoguelph.ca}}
}

\author{Matthew Kowal,
Mennatullah Siam, 
Md Amirul Islam \\
Neil D. B. Bruce,
Richard P. Wildes,
Konstantinos G. Derpanis \\
\IEEEcompsocitemizethanks{
\IEEEcompsocthanksitem M.Kowal, K.\ Derpanis and R.\ Wildes are with the Department
of Electrical Engineering and Computer Science, York University, Toronto.\\
M.\ Siam is with Engineering and Applied Science, Ontario Tech University and Computer Science, University of British Columbia.\\
Md A. Islam is with the Noah's Ark lab at Huawei Technologies Canada.\\
N. Bruce is with the Department of Computer Science, Guelph University.\\
E-mail: \{m2kowal, kosta, wildes\}@eecs.yorku.ca, mennatullah.siam@ubc.ca}
\thanks{Manuscript received July 30, 2022; }}

\IEEEtitleabstractindextext{%
\begin{abstract}
There is limited understanding of the information captured by deep spatiotemporal models in their intermediate representations. For example, while evidence suggests that action recognition algorithms are heavily influenced by visual appearance in single frames, no quantitative methodology exists for evaluating such static bias in the latent representation compared to bias toward dynamics. We tackle this challenge by proposing an approach for quantifying the static and dynamic biases of any spatiotemporal model, and apply our approach to three tasks, action recognition, automatic video object segmentation (AVOS) and video instance segmentation (VIS). Our key findings are: (i) Most examined models are biased toward static information. (ii) Some datasets that are assumed to be biased toward dynamics are actually biased toward static information. (iii) Individual channels in an architecture can be biased toward static, dynamic or jointly encode a combination static and dynamic information. (iv) Most models converge to their culminating biases in the first half of training. We then explore how these biases affect performance on dynamically biased datasets. For action recognition, we propose StaticDropout, a semantically guided dropout that debiases a model from static information toward dynamics. For AVOS, we design a better combination of fusion and cross connection layers compared with previous architectures.
\end{abstract}

}

\maketitle

\IEEEdisplaynontitleabstractindextext

%
\IEEEpeerreviewmaketitle


%
%
%
%

\section{Introduction}

\IEEEPARstart{T}{his} paper focuses on the problem of interpreting the information learned by deep neural networks (DNNs) trained for video understanding tasks. Interpreting deep spatiotemporal models is a largely understudied topic in computer vision despite their achieving state-of-the-art performance on video understanding tasks, such as action recognition~\cite{zhu2020comprehensive} and video object segmentation~\cite{wang2021survey}. These models are trained in an end-to-end fashion to learn discriminative static and dynamic features over space and time. 
Here, we use the term \textit{static} to refer to attributes that can be extracted from a single image (\eg color and texture) and the term \textit{dynamic} to attributes that arise from consideration of multiple frames (\eg motion and dynamic texture). 

While this learning-based paradigm has led to great success across a wide range of tasks, the internal representations of the learned models remain largely opaque. This lack of explainability is unsatisfying from both scientific and application perspectives. From a scientific perspective, there is limited understanding of what information is driving the decision-making underlying the network output. Elucidating
the decision-making process may yield directions to improve models. From an applications perspective, there have been multiple cases showing 
the ethical and damaging consequences of deploying opaque vision models, 
\eg~\cite{buolamwini2018gender,hansson2021self}.
Currently, however, the explainability of spatiotemporal models is under-explored~\cite{hiley2019explainable}. Some evidence suggests that these models exhibit considerable bias toward static information, \eg~\cite{vu2014predicting,he2016human,choi2019can,ilic2022appearance}; therefore, an interesting question 
to answer about 
the representations in deep spatiotemporal models is: \textit{How much static and dynamic information is being captured}? While a few video interpretation methods exist, they have various limitations, \eg being primarily qualitative~\cite{feichtenhofer2020deep}, using a certain dataset that prevents evaluating the effect of the training dataset~\cite{hadji2018new} or using classification accuracy as a metric without quantifying 
a model's \textit{internal} representations~\cite{hadji2018new, sevilla2021only,ilic2022appearance}.

\begin{figure}[t]
\centering
    \vspace{-0.5cm}
    \resizebox{0.48\textwidth}{!}{
    \includegraphics[]{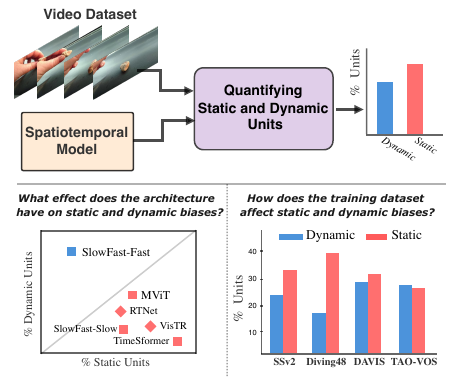}}
    \vspace{-0.5cm}
    \caption{We introduce a general technique that, given a trained spatiotemporal model and a video dataset, can quantify the bias in any hidden representation within the model toward encoding static (red) or dynamic (blue) information. We use this technique to study action recognition (squares) and video segmentation (diamonds) and explore the effect of models and training datasets on static and dynamic biases.}
\vspace{-3em}
\end{figure}\label{fig:motivation}

In response, we present a quantitative paradigm for evaluating the extent that spatiotemporal models are biased toward static or dynamic information in their internal representations.
We define bias toward a certain factor (dynamic or static) as the percentage of units (\ie channels) within intermediate layers that encode that factor; see Fig.~\ref{fig:motivation} (top). Inspired by previous work~\cite{esser2020disentangling,islam2021shape}, we propose a metric to estimate the amount of static \vs dynamic bias based on the mutual information between sampled video pairs corresponding to these factors. We explore three common tasks 
to show the efficacy of our approach as a general tool for understanding spatiotemporal models, action recognition, automatic video object segmentation (AVOS) and video instance segmentation (VIS). We focus our study on answering the following five questions: (i) How do model architectures affect static and dynamic biases? (ii) How does the training dataset affect these biases? (iii) What role do units that jointly encode static and dynamic information play in relation to the architecture and dataset? (iv) When are statics and dynamics learned during training? (v) What impact do these biases have on performance? In regards to the last question, we demonstrate that controlling the static and dynamic biases in action recognition and AVOS can improve performance on tasks requiring dynamics.

\noindent{\bf Contributions.} We make five main contributions: (i) We introduce a general method for quantifying the static and dynamic bias contained in spatiotemporal models, including a novel sampling procedure to produce static and dynamic video pairs. (ii) We propose a technique for identifying units that jointly encode static and dynamic factors. (iii) Using the aforementioned techniques, we provide a unified study on three widely researched tasks, action recognition, AVOS and VIS, with a focus on the effect of architecture and training dataset on a model's static and dynamic biases; see Fig.~\ref{fig:motivation} (bottom). (iv) We propose StaticDropout, a semantically guided dropout technique for debiasing models from statics and toward dynamics that can improve model performance on datasets which require dynamics. (v) We demonstrate how a proper selection of the fusion and cross connection modules in AVOS architectures can guide a model to learn better dynamics and improve performance for the task of segmenting camouflaged entities. This work extends our previous work~\cite{kowal2022deeper} by including a new task in our analysis (Video Instance Segmentation)~\ref{sec:vis_archs},~\ref{sec:vis_dataset} and~\ref{sec:epochwise_vis}, the effect of training on static and dynamic biases~\ref{sec:epoch_effect}, and introducing two methods for controlling the dynamic bias learned by action recognition~\ref{sec:staticdropout} and video object segmentation models~\ref{sec:vos_cc_study}.
We discover that all studied networks are heavily static biased, except for two-stream architectures with cross connections encouraging models to capture dynamics. We show that the majority of models converge to their final static and dynamic biases within the first half of training iterations. Additionally, we confirm that, contrary to previous beliefs~\cite{li2018resound,bertasius2021space}, the Diving48~\cite{li2018resound} dataset is not dynamically biased and Something-Something-v2 (SSv2)~\cite{goyal2017something} is better suited to evaluate a model's ability to capture dynamics. Publicly available code is available\footnote{\url{https://yorkucvil.github.io/Static-Dynamic-Interpretability/}}.

\section{Related work}




\noindent \textbf{Spatiotemporal models.} Deep spatiotemporal models that learn discriminative features across space and time have proven effective for 
video understanding tasks~\cite{aafaq2019video,zhu2020comprehensive}. 
Extant models 
can be broadly categorized (agnostic of the downstream task) into: two-stream approaches that separately model motion and appearance features~\cite{carreira2017quo,jain2017fusionseg,zhou2020motion,ren2021reciprocal,feichtenhofer2019slowfast}, 
3D convolutions that jointly model motion and appearance~\cite{carreira2017quo},
attention-based models with different forms of spatiotemporal data association~\cite{bertasius2021space,ren2021reciprocal}, models relying on recurrent neural networks~\cite{tokmakov2017learning} and hybrid models that combine elements of the aforementioned models~\cite{tokmakov2017learning,carreira2017quo, ren2021reciprocal}. 
Our approach to quantifying bias is not limited to the particulars of a model and is applicable to all extant and future models. We empirically demonstrate the flexibility of our approach by evaluating a diverse set of models.

\begin{figure*}[t]
    \includegraphics[width=\textwidth]{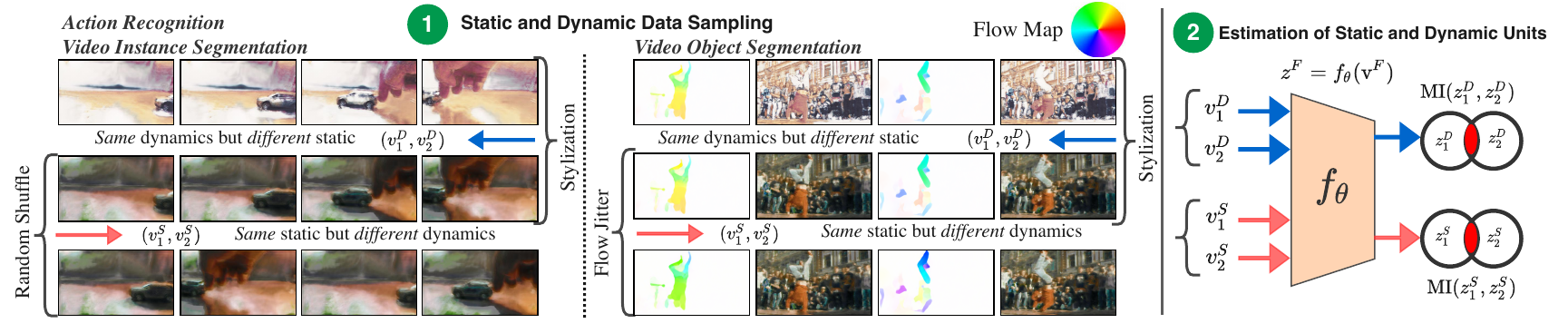}
    \vspace{-0.8cm}
    \caption{Overview of our methodology for analysing bias toward static or dynamic information. We measure the dynamic and static biases in deep spatiotemporal models for three tasks: action recognition, automatic video object segmentation and video instance segmentation. \textbf{(1)} We sample video pairs that share either \textit{static}, $(v^S_1, v^S_2)$, or \textit{dynamic}, ($v^D_1,v^D_2$), information using video stylization~\cite{texler2020interactive} and frame shuffling or optical flow jitter (flow visualized in RGB format). \textbf{(2)} Given a pretrained model, $f_\theta$, we compute the mutual information (MI) between intermediate representations of video pairs, $z^F$, to assess the model's bias toward either factor on a per-layer, $l$, or per-channel (\ie unit) basis. }
    \label{fig:mainmethod}
    \vspace{-0.6cm}
\end{figure*}

\noindent{\bf Action recognition.} 
3D convolutional networks are popular for learning 
spatiotemporal representations of videos for action recognition, \eg~\cite{taylor2010convolutional, ji20123d,tran2015learning,carreira2017quo,hara2017learning}. 
Other work has considered two-stream architectures,
where the dynamics were provided directly to one of the streams as optical flow, \eg~\cite{simonyan2014two,feichtenhofer2017spatiotemporal}.
Representative of the state-of-the-art with convolutional networks is SlowFast~\cite{feichtenhofer2019slowfast}, which
is a two-stream 3D CNN that only takes RGB videos as input.
To encourage each stream to specialize in capturing predominately static or dynamic information, 
the temporal sampling rates of the inputs to each stream differ.
Recently, attention based approaches have proven to be suited to both static and time-series visual data, including action recognition, with variants of the transformer architecture~\cite{vaswani2017attention,bertasius2021space,fan2021multiscale,patrick2021keeping}.


\noindent{\bf Video segmentation.} Deep video segmentation approaches can be categorized into two categories~\cite{wang2021survey}: (i) class agnostic, which are referred to as video object segmentation (VOS) and (ii) video semantic segmentation, which predict different semantic categories. Video semantic segmentation has been investigated with both instance-agnostic~\cite{gadde2017semantic} and instance-aware~\cite{wang2021end} approaches, where the latter is referred to as video instance segmentation (VIS)~\cite{wang2021end}. Video object segmentation methods can be categorized into automatic, semi-automatic and interactive~\cite{wang2021survey}. AVOS tries to segment the visual and motion salient objects in videos without a predefined mask initialization, while semi-automatic VOS requires mask initialization in the first frame to consequently track objects within a video. In this work, we mainly focus on AVOS approaches that segment salient objects in videos. 
We consider two-stream models that fuse motion and appearance features. We also investigate the effect of no cross connections~\cite{jain2017fusionseg} relative to both motion-to-appearance~\cite{zhou2020motion} or bidirectional~\cite{ren2021reciprocal} cross connections. Additionally, we explore video instance segmentation approaches that use raw images as input to demonstrate the versatility of our methodology across different video segmentation tasks.

\noindent \textbf{Interpretability of spatiotemporal models.}
Limited work has been dedicated to the interpretability of spatiotemporal models. Several efforts predicate model interpretation on proxy tasks, \eg dynamic texture recognition~\cite{hadji2018new} or future frame selection~\cite{ghodrati2018video}. These approaches do not interpret the learned representations in the intermediate layers and in some cases require training to be performed on specific datasets~\cite{hadji2018new}. Work also has appeared that introduced a dataset to completely decouple static and dynamic information, but used it only to examine overall architecture performance on action recognition and did not examine intermediate representations~\cite{ilic2022appearance}. Other work focused on understanding latent representations in spatiotemporal models either mostly concerned qualitative visualization~\cite{feichtenhofer2020deep} or a specific architecture type~\cite{zhao2021interpretable}. A related task is understanding the scene representation bias of action recognition datasets~\cite{li2018resound, li2019repair}. 
Recently, a method was proposed to discover spatiotemporal concepts in video transformers~\cite{kowal2024understanding}.
However, these efforts did not focus on the effect of different architectural inductive biases on the learned intermediate representations. Our proposed interpretability technique is
the first to \emph{quantify} static and dynamic biases on \textit{intermediate} representations learned in off-the-shelf models for multiple video-based tasks. Most prior efforts focused on a single task, and studied either datasets~\cite{li2018resound} or architectures~\cite{feichtenhofer2020deep,manttari2020interpreting}. In contrast, our unified study covers seven datasets and dozens of architectures on three different tasks, \ie action recognition, AVOS and VIS. 

\section{Methodology}
We introduce an approach to quantify the number of units encoding static and dynamic information in spatiotemporal models; for an overview, see 
Fig.~\ref{fig:mainmethod}. Our approach consists of two main steps. First, given a pretrained spatiotemporal model, we sample static and dynamic video pairs (Sec.~\ref{sec:sampling}). 
Second, we use the static and dynamic pairs to estimate the number of units in the model encoding each factor based on the mutual information shared between the pairs (Sec.~\ref{sec:mmi}).

\vspace{-0.4cm}
\subsection{Sampling static and dynamic pairs}\label{sec:sampling}
\noindent \textbf{Why static and dynamic?}
We define static as `information arising from single frames' and dynamic as `information arising from the consideration of multiple frames'. The main alternative attribute to dynamics that we considered is `image motion' (\ie trackable points or regions), but `motion' is a subset of dynamic information~\cite{wildes2000qualitative,derpanis2011spacetime} (\eg stationary flashing lights have dynamics but no motion). Thus, we consider dynamics over motion because it encompasses a wider range of visual phenomena. In complement, we choose the term `static' over the possible alternative `appearance', because dynamics also can provide appearance information,~\eg object contours, even if camouflaged in a single frame, can be revealed through motion. 
For our metric, we produce video pairs that contain the same static information and perturbed dynamics, or vice versa, with the end goal of analyzing models trained on large-scale real-world datasets. The dynamic perturbation is done via frame shuffling or optical flow jitter (depending on the model input), while the static perturbation is achieved via video stylization. We note that \textit{shape} is a confounder between the static and dynamic factors: It is not feasible to completely disentangle static and dynamic as perceiving the motion of an object necessarily provides localized boundary information of that object. Despite this confounder, video stylization still provides notable differences as there is less mutual appearance information post-stylization (Sec.~\ref{sec:experiments}). We now detail our static and dynamic sampling techniques, as visualized in Fig.~\ref{fig:mainmethod} (panel 1).

\noindent \textbf{Action recognition.} The action recognition models we consider take in multiple frames (four to thirty-two). To construct video pairs with the \textit{same} dynamics but \textit{different} static information (\ie~\textit{dynamic pairs}), we consider the same video but with two \textit{different} video styles. For video stylization, we use a recent video stylization method (with four possible styles) that perturbs static attributes like color, pixel intensity and texture~\cite{texler2020interactive}, but has less temporal artifacts (\eg flickering) than stylization methods that consider each image independently~\cite{huang2017arbitrary}. These video pairs will contain objects and scenes that have identical dynamics, but have perturbed static information. To construct pairs with the \textit{same} static information but \textit{different} dynamics (\ie~\textit{static pairs}), we take two videos of the same style, but randomly \textit{shuffle} the frames along the temporal axis; see Fig.~\ref{fig:mainmethod} (panel 1, left). In this case, the temporal correlations are altered while the static (\ie per-frame) information remains identical.

\noindent \textbf{Video object segmentation.} The AVOS models considered~\cite{jain2017fusionseg,zhou2020motion,ren2021reciprocal} take a single RGB frame and an optical flow frame as input to the appearance and motion streams, resp.; 
see Fig.~\ref{fig:mainmethod} (panel 1, right). Therefore, we apply an alternative method to frame shuffling to obtain the \textit{static} pairs. For the \textit{static} pair, we use RGB images with the \textit{same} style but alter the dynamics by jittering the optical flow. 
To do this, we represent flow with a color coding~\cite{baker2011database} and then randomly perturb the hue and saturation which correspond to the direction and magnitude, respectively.
For the \textit{dynamic} pairs, we use the \textit{same} optical flow but a \textit{different} image style. For creating stylized images, we use the same video stylization techniques noted above for action recognition~\cite{texler2020interactive}, and then sample frames from the generated video. 

\noindent \textbf{Video instance segmentation.} The inputs to the VIS models considered~\cite{wang2021end} take single stream, multi-frame RGB inputs. Therefore, we select static and dynamic pairs similar to action recognition models, \ie shuffling and stylization for static and dynamic pairs, respectively.


\vspace{-0.3cm}
\subsection{Estimating static and dynamic units} \label{sec:mmi}

We seek to quantify the number of units (\ie \textit{channels}) in a layer encoding \textit{static} or \textit{dynamic} information and the extent to which individual units perform static, dynamic or joint encodings. Inspired by recent work that focused on single images~\cite{esser2020disentangling,islam2021shape}, we use 
a mutual information estimator to measure the information shared between video pairs. 

\noindent\textbf{Layer-wise metric.} Given a pre-trained network, $f_\theta$, and a pair of videos, $v^F_1$
and $v^F_2$, that share the semantic factor $F$ (\ie \textit{static} or \textit{dynamic}), we compute the features for an intermediate layer $l$ as $z^F_1 = f^l_{\theta}(v^F_1)$ and $z^F_2 = f^l_{\theta}(v^F_2)$ (omitting the $l$ on $z$ to reduce notation). We use $z^F_1(i), z^F_2(i)$ to denote the $i^{\text{th}}$ unit (\ie channel) in $N^l$ dimensional features after a global average pooling layer. Units biased toward the \textit{static} factor will result in a higher correlation among \textit{static} pairs than the \textit{dynamic} pairs and vice versa. Under the assumption that units in the intermediate representation, $z^F_1(i)$ and $z^F_2(i)$, across the dataset are jointly Gaussian, the correlation coefficient can be used as a lower bound on mutual information~\cite{kraskov2004estimating,foster2011lower}, as used in previous work~\cite{esser2020disentangling,islam2021shape}.
The number of units encoding factor $F$, $N_F$, is obtained by computing the correlation coefficient, $S_F$, over all $N^l$ channels between all video pairs $z^F_1, z^F_2$, as
\vspace{-0.3cm}
\begin{equation}
\begin{split}
  N_F = \frac{\exp{(S_F)}}{\sum\limits_{k=0}^K{\exp{(S_k)}}} \cdot N^l,
     S_F = \sum\limits_{i=1}^{N^l} \frac{\text{Cov}(z^F_1(i), z^F_2(i))}{
        \sqrt{\text{Var}(z^F_1(i)) \;\text{Var}(z^F_2(i))}},
\end{split}\label{eq:biasscores}
\end{equation}
where we multiply the Softmax by the number of units in that layer, $N^l$, to compute the number of units encoding the semantic factor $F$ relative to the other factors considered and $K=\{{\text{static}, \text{dynamic}, \text{identical}}\}$. In addition to \textit{static} and \textit{dynamic}, we consider a third factor in (\ref{eq:biasscores}), the \textit{identical} factor, where the video pairs have the same static and dynamic factors (\ie same video, style, frame ordering and optical flow). This baseline factor is the correlation between the model's encoding of the same videos, that gives $S_\text{Identical}=1$ for all layers. 


\noindent\textbf{Unit-wise metric.} The correlation coefficient, $S_F$, estimates the relative amount of static and dynamic information over all units in a particular layer; note the pooling done by the summation \textit{before} the Softmax in the layer-wise metric, (\ref{eq:biasscores}). However, it is also desirable to measure static and dynamic information contained in each individual channel. This measurement allows for a more fine-grained analysis of how many channels (\ie units) encode a factor $F$ above a certain threshold, as well as identify any joint or residual (\ie non-dynamic or static) units. Thus, we categorize each unit based on how much information (\ie static \vs dynamic) is encoded, whether any units jointly encode both factors or if there are units that do not correlate with either type of information. We measure the amount of static and dynamic information encoded in each unit  $i \in {1, \dotsc, N^l}$ as
\begin{equation}
s^i_{F} = \frac{\text{Covariance}(z^F_1(i), z^F_2(i))}{\sqrt{ \text{Variance}( z^F_1(i)) \text{Variance}( z^F_2(i))}},\label{eq:bias_scores_indv}
\end{equation} 
where each $s^i_{F}$ is the information of semantic factor $F$ in unit $i$. Given these individual correlations, we calculate the individual factors by excluding the use of a Softmax and simply threshold the correlation for each factor with a constant parameter, $\lambda$, to yield our unit-wise metrics as
\begin{equation}
\begin{split}
N_{\text{F}} = \sum_{i=1}^{N^l}\mathbbm{1}[ s^i_{F} > \lambda \land s^i_{k} < \lambda \forall k \in K, k \neq F] \\
N_{\text{J}} = \sum_{i=1}^{N^l}\mathbbm{1}[ s^i_{F} > \lambda \forall F \in K], \,
N_{\text{R}} = \sum_{i=1}^{N^l}\mathbbm{1}[ s^i_F < \lambda \forall F \in K],
\end{split}
\label{eq:ind_bias_scores_diff_b}
\end{equation}
where $K=\{\text{static}, \text{dynamic}\}$, $N_{\text{J}}$ indicates units jointly encoding both and $N_{\text{R}}$ are residual units not correlating with these factors under a threshold, $\lambda$. Note that we assign units to either joint, dynamic, static or residual and do not allow for an overlap to occur. This approach allows us to investigate the existence of units that jointly encode static and dynamic factors. For all experiments, we set $\lambda=0.5$ since it is halfway between \textit{no} and \textit{full} positive correlation. 

We note that the term `bias' also has been used to describe the tendency of a model to make predictions with specific qualities (e.g., shape vs. texture labels~\cite{geirhos2018imagenet}). In our work, we define `bias' towards a specific factor (i.e., static or dynamic) to be the layer (or unit) encoding more relative mutual information of that factor than another (as used in~\cite{islam2021shape}).

\subsection{Model Biasing}
Following an analysis of static and dynamic biases in various architectures, tasks, and datasets, we then aim to control a model's bias to learn dynamic information for two different tasks in Sec.~\ref{sec:app}. First, we introduce StaticDropout, a new dropout mechanism that encourages action recognition classification models to encode more dynamic information. More specifically, we use Eq.~\ref{eq:bias_scores_indv} to calculate the channels in an action recognition model that encode static information, and drop them out every few iterations over the course of training. We show that increasing the number of units dropped corresponds with more dynamic information being encoded in the model.

For video object segmentation, we provide a formal analysis on the different types of fusion and cross connection layers found in two-stream models. We propose a simple combination of layers that we empirically show to encode more dynamic information, using Eq.~\ref{eq:bias_scores_indv}. Moreover, we show this simple improvement can increase model performance on dynamic centric tasks (e.g., camouflaged object segmentation~\cite{lamdouar2020betrayed}) by up to 10\% mean Intersection over Union (mIoU).

\section{Empirical results}\label{sec:experiments}
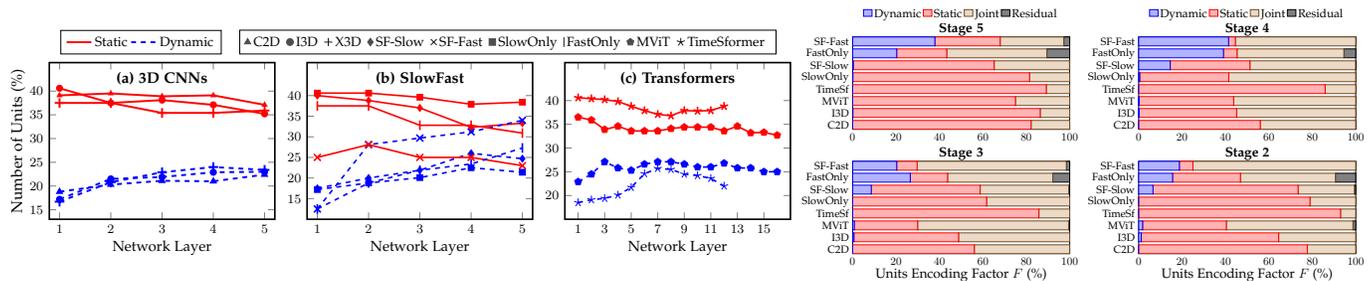
\begin{figure*} [t]
	\begin{center}
     \centering
\begin{minipage}{0.58\textwidth}
		\resizebox{\textwidth}{!}{
\begin{tikzpicture}\ref{legend_all} \ref{legend_color}
    \begin{groupplot}[group style = {group size = 3 by 1, horizontal sep = 20pt}, width = 6.0cm, height = 6.4cm]
\nextgroupplot[
      line width=1.0,
        title={\textbf{(a) 3D CNNs}},
        title style={at={(axis description cs:0.5,0.92)},anchor=north},
        xlabel={Network Layer},
        ylabel={Number of Units (\%)},
        xmin=0.8, xmax=5.2,
        ymin=13, ymax=46,
        xtick={1,2,3,4,5},
        ytick={15, 20, 25, 30, 35, 40},
        x tick label style={font=\footnotesize},
        y tick label style={font=\footnotesize},
        x label style={at={(axis description cs:0.5,0.06)},anchor=north,font=\small}, 
        y label style={at={(axis description cs:0.17,.5)},anchor=south,font=\normalsize},
        width=6.5cm,
        height=5cm,        
        ymajorgrids=false,
        xmajorgrids=false,
        major grid style={dotted,green!20!black},
    ]
    \addplot[line width=1.2pt,dashed,mark options={scale=0.8,solid},color=blue!100,mark=triangle*,]
        coordinates {(1,18.8)(2,20.3)(3,21.1)(4,21.0)(5,22.4)};
    \addplot[line width=1.2pt,mark size=1.1pt,color=red!100,mark=triangle*,]
        coordinates {(1,39.1)(2,39.5)(3,38.9)(4,39.1)(5,37.1)};
    \addplot[line width=1.2pt,dashed,mark options={scale=0.8,solid},color=blue!100,mark=*,]
        coordinates {(1,17.2)(2,21.5)(3,21.9)(4,22.9)(5,23.0)};
    \addplot[line width=1.2pt,mark options={scale=0.8,solid},color=red!100,mark=*,]
        coordinates {(1,40.6)(2,37.5)(3,38.1)(4,37.1)(5,35.2)};
    \addplot[line width=1.2pt,dashed,mark options={scale=1.5,solid},color=blue!100,mark=+,]
        coordinates {(1,16.7)(2,20.8)(3,22.9)(4,24.0)(5,23.4)};
    \addplot[line width=1.2pt,mark options={scale=1.5,solid},color=red!100,mark=+,]
        coordinates {(1,37.5)(2,37.5)(3,35.4)(4,35.4)(5,35.9)};

\nextgroupplot[
      line width=1.0,
        title={\textbf{(b) SlowFast}},
        title style={at={(axis description cs:0.5,0.92)},anchor=north,font=\normalsize},
        xlabel={Network Layer},
        xmin=0.8, xmax=5.2,
        ymin=10, ymax=48,
        xtick={1,2,3,4,5},
        ytick={15, 20, 25, 30, 35, 40},
        x tick label style={font=\footnotesize},
        y tick label style={font=\footnotesize},
        x label style={at={(axis description cs:0.5,0.06)},anchor=north,font=\small},   
        width=6.5cm,
        height=5cm,        
        ymajorgrids=false,
        xmajorgrids=false,
        major grid style={dotted,green!20!black},
        legend style={
        nodes={scale=0.87, transform shape},
        cells={anchor=west},
        legend style={at={(2,3.6)},anchor=south,row sep=0.01pt}, font =\normalsize},
        legend image post style={scale=0.9},
        legend columns=2,
        legend to name=legend_color,
    ]
    \addplot[line width=1pt,dashed,mark options={scale=0.9,solid},color=blue!100,mark=diamond*,forget plot]
        coordinates {(1,17.5)(2,20.0)(3,21.9)(4,26.0)(5,24.7)};
    \addplot[line width=1pt,mark options={scale=0.9,solid},color=red!100,mark=diamond*,forget plot]
        coordinates {(1,40.0)(2,38.8)(3,37.0)(4,32.3)(5,33.3)};
    \addplot[line width=1pt,dashed,mark options={scale=1.5,solid},color=blue!100,mark=x,forget plot]
        coordinates {(1,12.5)(2,28.1)(3,29.7)(4,31.2)(5,34.0)};
    \addplot[line width=1pt,mark options={scale=1.5,solid},color=red!100,mark=x,forget plot]
        coordinates {(1,25.0)(2,28.1)(3,25.0)(4,25.0)(5,23.0)};
    \addplot[line width=1pt,dashed,mark options={scale=0.8,solid},color=blue!100,mark=square*,forget plot]
        coordinates {(1,17.2)(2,18.8)(3,20.1)(4,22.5)(5,21.4)};
    \addplot[line width=1pt,mark options={scale=0.8,solid},color=red!100,mark=square*,forget plot]
        coordinates {(1,40.6)(2,40.6)(3,39.6)(4,37.9)(5,38.4)};
    \addplot[line width=1pt,dashed,mark options={scale=1.5,solid},color=blue!100,mark=|,forget plot]
        coordinates {(1,12.5)(2,18.8)(3,21.9)(4,23.4)(5,27.3)};
    \addplot[line width=1pt,mark options={scale=1.5,solid},color=red!100,mark=|,forget plot]
        coordinates {(1,37.5)(2,37.5)(3,32.8)(4,32.8)(5,30.9)};
    
    \addlegendimage{line width=1.2pt,color=red}\label{pgfplots:ar_stat}
    \addlegendentry[color=black]{Static}
    \addlegendimage{line width=1.2pt,dashed,color=blue}\label{pgfplots:ar_dyn}
    \addlegendentry[color=black]{Dynamic}
          
\nextgroupplot[
      line width=1.0,
        title={\textbf{(c) Transformers}},
        title style={at={(axis description cs:0.5,0.92)},anchor=north,font=\normalsize},
        xlabel={Network Layer},
        xmin=0, xmax=17,
        ymin=15, ymax=48,
        xtick={1,3,5,7,9,11,13,15},
        ytick={20, 25, 30, 35, 40},
        x tick label style={font=\footnotesize},
        y tick label style={font=\footnotesize},
        x label style={at={(axis description cs:0.5,0.06)},anchor=north,font=\small},
        width=6.5cm,
        height=5cm,         
        ymajorgrids=false,
        xmajorgrids=false,
        major grid style={dotted,green!20!black},
        legend style={
         legend style={row sep=0.1pt},
        nodes={scale=0.87, transform shape},
        legend columns=-1,
        cells={anchor=west},
        legend style={at={(10,3.6)},anchor=south,row sep=0.01pt}, font =\normalsize},
        legend to name=legend_all,
    ]
    \addplot[line width=1pt,dashed,mark options={scale=1,solid},color=blue!100,mark=pentagon*,forget plot]
        coordinates {(1,22.9)(2,24.5)(3,27.1)(4,25.8)(5,25.3)(6,26.6)(7,27.1)(8,27.1)(9,26.6)(10,26.0)(11,26.0)(12,26.8)(13,25.8)(14,25.8)(15,25.0)(16,25.0)};
    \addplot[line width=1pt,mark options={scale=1,solid},color=red!100,mark=pentagon*,forget plot]
        coordinates {(1,36.5)(2,35.9)(3,33.9)(4,34.6)(5,33.6)(6,33.6)(7,33.6)(8,34.1)(9,34.4)(10,34.4)(11,34.4)(12,33.6)(13,34.6)(14,33.2)(15,33.3)(16,32.7)};
    \addplot[line width=1pt,dashed,mark options={line width=0.5pt,scale=1.3,solid},color=blue!100,mark=star,forget plot]
        coordinates {(1,18.5)(2,19.1)(3,19.4)(4,20.1)(5,21.7)(6,24.6)(7,25.7)(8,25.5)(9,24.5)(10,24.2)(11,23.6)(12,22.0)};
    \addplot[line width=1pt,mark options={scale=1.3,solid},color=red!100,mark=star,forget plot]
        coordinates {(1,40.6)(2,40.4)(3,40.2)(4,39.8)(5,38.8)(6,37.9)(7,37.1)(8,36.8)(9,37.9)(10,37.8)(11,37.9)(12,38.8)};

    
    \addlegendimage{only marks,mark=triangle*,mark size=2.2pt,color=black!70}\label{pgfplots:ar_c1r1}
\addlegendentry[color=black]{C2D  \hspace{1cm}}
    \addlegendimage{only marks,mark=*,mark size=2.1pt,color=black!70}\label{pgfplots:ar_c1r2}
\addlegendentry[color=black]{I3D  \hspace{1cm}}
    \addlegendimage{only marks,mark=+,mark size=2.5pt,color=black!90}\label{pgfplots:ar_c1r3}
\addlegendentry[color=black]{X3D  \hspace{1cm}}
    \addlegendimage{only marks,mark=diamond*,mark size=2.3pt,color=black!70}\label{pgfplots:ar_c1r4}
\addlegendentry[color=black]{SF-Slow  \hspace{1cm}}
    \addlegendimage{only marks,mark=x,mark size=2.6pt,color=black!90}\label{pgfplots:ar_c1r5}
\addlegendentry[color=black]{SF-Fast  \hspace{1cm}}
    \addlegendimage{only marks,mark=square*,mark size=2pt,color=black!70}\label{pgfplots:ar_c1r6}
\addlegendentry[color=black]{SlowOnly  \hspace{1cm}}
    \addlegendimage{only marks,mark=|,mark size=3pt,color=black!90}\label{pgfplots:ar_c1r7}
\addlegendentry[color=black]{FastOnly  \hspace{1cm}}
    \addlegendimage{only marks,mark=pentagon*,mark size=2.2pt,color=black!70}\label{pgfplots:ar_c1r8}
\addlegendentry[color=black]{MViT  \hspace{1cm}}
    \addlegendimage{only marks,mark=star,mark size=2.6pt,color=black!90}\label{pgfplots:ar_c1r9}
\addlegendentry[color=black]{TimeSformer  \hspace{1cm}}

\end{groupplot}

    
\end{tikzpicture}
}
\end{minipage}%
\begin{minipage}{0.42\textwidth}
\resizebox{1.0\textwidth}{!}{

\begin{tikzpicture}
\begin{axis} [xbar stacked,
    width=\axisdefaultwidth,
    height=4.5cm,
    bar width = 8pt,
    xmin = 0,
    xmax = 100,
    title = \textbf{Stage 5},
    title style={at={(axis description cs:0.5,1.12)},anchor=north,font=\large},
    ytick=data,
    legend style={
         draw=none,
         legend style={row sep=0.1pt},
        nodes={scale=0.87, transform shape},
        legend columns=-1,
        cells={anchor=west},
        legend style={at={(0.5,1.35)},anchor=north,row sep=0.01pt}, font =\large},
    symbolic y coords={C2D, I3D, MViT, TimeSf, SlowOnly, SF-Slow, FastOnly, SF-Fast},
    enlarge x limits = {value = .1},
    enlarge y limits={abs=4pt}
]

\addplot coordinates {(0,C2D) [0] (0.1,I3D) [2] (0,MViT) [0] (0,TimeSf) [0] (0.1,SlowOnly) [2] (0.3,SF-Slow) [6] (20.3125,FastOnly) [23] (37.890625,SF-Fast) [101]}; 
\addplot coordinates {(82.2,C2D) [1684] (86.3,I3D) [1767] (75,MViT) [576] (89.2,TimeSf) [685] (81.4,SlowOnly) [1668] (64.9,SF-Slow) [1329] (23.046875,FastOnly) [82] (30.078125,SF-Fast) [72]}; 
\addplot coordinates {(17.8,C2D) [364] (13.6,I3D) [278] (25,MViT) [192] (10.8,TimeSf) [83] (18.5,SlowOnly) [378] (34.7,SF-Slow) [710] (46.09375,FastOnly) [149] (29.296875,SF-Fast) [78]}; 
\addplot coordinates {(0,C2D) [0] (0.05,I3D) [1] (0,MViT) [0] (0,TimeSf) [0] (0,SlowOnly) [0] (0.14,SF-Slow) [3] (10.546875,FastOnly) [2] (2.734375,SF-Fast) [5]}; 


\legend {Dynamic,Static,Joint,Residual};

\end{axis}
\end{tikzpicture} 
\begin{tikzpicture}
\begin{axis} [xbar stacked,
    width=\axisdefaultwidth,
    height=4.5cm,
    bar width = 8pt,
    xmin = 0,
    xmax = 100,
    title = \textbf{Stage 4},
    title style={at={(axis description cs:0.5,1.12)},anchor=north,font=\large},
    ytick=data,
    legend style={
         draw=none,
         legend style={row sep=0.1pt},
        nodes={scale=0.87, transform shape},
        legend columns=-1,
        cells={anchor=west},
        legend style={at={(0.5,1.35)},anchor=north,row sep=0.01pt}, font =\large},
    symbolic y coords={C2D, I3D, MViT, TimeSf, SlowOnly, SF-Slow, FastOnly, SF-Fast},
    enlarge x limits = {value = .1},
    enlarge y limits={abs=4pt}
]

\addplot coordinates {(0,C2D) (0.1,I3D) (0.48828125,SlowOnly) (0,TimeSf) (14.53125,SF-Slow) (41.40625,SF-Fast) (39.0625,FastOnly) (0.13020833333333334,MViT)};
\addplot coordinates {(56,C2D) (45.02,I3D) (41.015625,SlowOnly) (85.9,TimeSf) (36.71875,SF-Slow)  (3.125,SF-Fast) (6.25,FastOnly) (43.619791666666664,MViT)};
\addplot coordinates {(44,C2D) (54.88,I3D) (58.49609375,SlowOnly) (14.1,TimeSf) (48.671875,SF-Slow) (55.46875,SF-Fast) (49.21875,FastOnly) (56.25,MViT)};
\addplot coordinates {(0,C2D) (0,I3D) (0.0,SlowOnly) (0,TimeSf) (0.078125,SF-Slow) (0.0,SF-Fast) (5.46875,FastOnly) (0.0,MViT)};
\legend {Dynamic,Static,Joint,Residual};

\end{axis}
\end{tikzpicture}
}
\vfill
\resizebox{1.0\textwidth}{!}{
\begin{tikzpicture}
\begin{axis} [xbar stacked,
    width=\axisdefaultwidth,
    height=4.5cm,
    bar width = 8pt,
    xmin = 0,
    xmax = 100,
    title = \textbf{Stage 3},
    title style={at={(axis description cs:0.5,1.12)},anchor=north,font=\large},
    xlabel = Units Encoding Factor $F$ (\%),
    ytick=data,
    xlabel style = {font=\large,at={(axis description cs:0.5,0.07)}},
    legend style={
			area legend,
			at={(0.5,1)},
			anchor=north,
			legend columns=-1},
    symbolic y coords={C2D, I3D, MViT, TimeSf, SlowOnly, SF-Slow, FastOnly, SF-Fast},
    enlarge x limits = {value = .1},
    enlarge y limits={abs=4pt}
]

\addplot coordinates {(0,C2D) (0.6,I3D) (0.0,SlowOnly) (0,TimeSf) (8.59375,SF-Slow) (20.3125,SF-Fast) (26.5625,FastOnly) (0.78125,MViT)};
\addplot coordinates {(56.05,C2D) (48.2,I3D) (61.71875,SlowOnly) (85.8,TimeSf) (50.15625,SF-Slow) (9.375,SF-Fast) (17.1875,FastOnly) (29.166666666666668,MViT)};
\addplot coordinates {(43.9,C2D) (51.2,I3D) (38.28125,SlowOnly) (14.2,TimeSf) (40.78125,SF-Slow) (68.75,SF-Fast) (48.4375,FastOnly) (69.53125,MViT)};
\addplot coordinates {(0,C2D) (0,I3D) (0.0,SlowOnly) (0,TimeSf) (0.46875,SF-Slow) (1.5625,SF-Fast) (7.8125,FastOnly) (0.5208333333333334,MViT)};


\end{axis}
\end{tikzpicture}
\begin{tikzpicture}
\begin{axis} [xbar stacked,
    width=\axisdefaultwidth,
    height=4.5cm,
    bar width = 8pt,
    xmin = 0,
    xmax = 100,
    title = \textbf{Stage 2},
    title style={at={(axis description cs:0.5,1.12)},anchor=north,font=\large},
    xlabel = Units Encoding Factor $F$ (\%),
    ytick=data,
    xlabel style = {font=\large,at={(axis description cs:0.5,0.07)}},
    legend style={
			area legend,
			at={(0.5,1)},
			anchor=north,
			legend columns=-1},
    symbolic y coords={C2D, I3D, MViT, TimeSf, SlowOnly, SF-Slow, FastOnly, SF-Fast},
    enlarge x limits = {value = .1},
    enlarge y limits={abs=4pt}
]

\addplot coordinates {(0,C2D) (1.2,I3D) (0.0,SlowOnly) (0,TimeSf) (6.5625,SF-Slow) (18.75,SF-Fast) (15.625,FastOnly) (1.8229166666666667,MViT)};
\addplot coordinates {(77.7,C2D) (63.3,I3D) (78.90625,SlowOnly) (93,TimeSf) (66.875,SF-Slow) (6.25,SF-Fast) (31.25,FastOnly) (38.541666666666664,MViT)};
\addplot coordinates {(22.3,C2D) (35.5,I3D) (21.09375,SlowOnly) (7,TimeSf) (25.9375,SF-Slow) (75.0,SF-Fast) (43.75,FastOnly) (58.333333333333336,MViT)};
\addplot coordinates {(0,C2D) (0,I3D) (0.0,SlowOnly) (0,TimeSf) (0.625,SF-Slow) (0.0,SF-Fast) (9.375,FastOnly) (1.3020833333333333,MViT)};

\end{axis}
\end{tikzpicture}
}
\end{minipage}%
	\end{center}
	\vspace{-0.5cm}
	\caption{Layer and unit-wise analyses on action recognition networks trained on Kinetics-400~\cite{carreira2017quo}. \textbf{Left:} Layer-wise encoding of static and dynamic factors using the layer-wise metric, (Eq.~\ref{eq:biasscores}), for: (a) single stream 3D CNNs, (b) SlowFast variants and (c) transformer variants. SF-Slow and SF-Fast denote the representation taken before the fusion layer from the slow and fast branches, resp. \textbf{Right:} Estimates of the dynamic, static, joint and residual units using the unit-wise metric, (Eq.~\ref{eq:ind_bias_scores_diff_b}).
	}\label{fig:stagewise_ar_all_models}
	\vspace{-0.3cm}
\end{figure*}

We choose the tasks of action recognition, AVOS, and VIS to demonstrate the generality of our approach. They differ in their semantics (\ie multi-class \vs binary \vs instance-based), labelling (\ie video-level \vs pixel-level), and input types (multi-frame images \vs single frame and optical flow). We explore five main research questions and show the corresponding results with respect to our quantitative techniques for both tasks: (i) What is the effect of the model architecture on the \textit{static} and \textit{dynamic} biases (Sec.~\ref{sec:architectures})? (ii) What effect does the training dataset have on \textit{static} and \textit{dynamic} biases (Sec.~\ref{sec:dataset_effect})? (iii) What are the characteristics of jointly encoding units in relation to model architectures and datasets (Sec.~\ref{sec:architectures} and~\ref{sec:dataset_effect})?  (iv) When are static and dynamic biases learned during training (Sec.~\ref{sec:epoch_effect})? (v) What is the effect of static and dynamic biased units on performance (Sec.~\ref{sec:neuron_removal})? Finally, we demonstrate two approaches where we use our insights to improve the performance of action recognition (Sec.~\ref{sec:staticdropout}) and AVOS models (Sec.~\ref{sec:vos_cc_study}).


\vspace{-0.3cm}
\subsection{Model architectures}
\label{sec:architectures}

\subsubsection{Action recognition}\label{sec:ar_models}
\noindent\textbf{Architectures.} 
We study three types of models with respect to their static and dynamic biases: (i) single stream 3D CNNs (\ie C2D~\cite{wang2018non}, I3D~\cite{carreira2017quo} and X3D~\cite{feichtenhofer2020x3d} models), (ii) SlowFast~\cite{feichtenhofer2019slowfast} variations, where the individual streams are referred to as the SlowOnly and FastOnly models and (iii) transformer-based architectures~\cite{fan2021multiscale,bertasius2021space}. For all models, the number of frames and sampling rate is ($8\times8$), except for the FastOnly network ($32\times2$), MViT ($16\times4$), and TimeSformer ($8\times32$). To identify the static and dynamic units of all models, we generate the Stylized ActivityNet~\cite{caba2015activitynet} validation set and use it for sampling \textit{static} and \textit{dynamic} pairs. We choose this dataset since the action distribution is similar to Kinetics-400, yet much smaller in size making it memory efficient when computing our metrics.

\noindent\textbf{Implementation details.} All models were taken from the SlowFast~\cite{feichtenhofer2019slowfast} repository\footnote{\url{https://github.com/facebookresearch/SlowFast}} except the TimeSformer~\cite{bertasius2021space}, which has its own codebase\footnote{\url{https://github.com/facebookresearch/TimeSformer}}. All model weights trained on Kinetics-400~\cite{carreira2017quo} and Something-Something-v2~\cite{goyal2017something} (SSv2) are taken directly from the SlowFast repository, except for the FastOnly model which we train ourselves. All training strategies are chosen to be similar to the original ones found in the SlowFast repository. We use decaying learning rate protocols to ensure that models have fully converged. The FastOnly model is trained on Kinetics-400 for 40 epochs with SGD, a weight decay of 1e-4, a batch size of 32 and a base learning rate of 0.03 that decreases by a factor of 10 at epochs 15, 30 and 35. On SSv2, the FastOnly model is trained for 25 epochs with SGD, weight decay of 1e-4, a batch size of 32 and a base learning rate that is decreased by a factor of 10 at epochs 10 and 20. On Diving48~\cite{li2018resound} the FastOnly model is trained for 100 epochs with SGD, weight decay of 1e-4, a batch size of 32 and a base learning rate of 0.0375 that decreases by a factor of 10 at epochs 40, 60 and 80. The SlowOnly model is trained on Diving48 for 100 epochs with SGD, weight decay of 1e-4, a batch size of 32 and a base learning rate of 0.00375 that decreases by a factor of 10 at epochs 40, 60 and 80. All models trained with temporal frame shuffling (see Sec.~\ref{sec:ar_dataset}) incur the same hyperparameters as their unshuffled counterparts.

\noindent \textbf{Layer-wise analysis.} 
Figure~\ref{fig:stagewise_ar_all_models} (left) shows the layer-wise analysis, (Eq.~\ref{eq:biasscores}), for various action recognition models. The transformers are measured at every layer and the convolutional architectures are measured at five `stages', corresponding to ResNet blocks~\cite{he2016deep}. Interestingly, all single stream networks are biased toward static information at all layers even though the video frames of the static pairs are randomly shuffled. Most of the 3D CNNs (\eg I3D and SlowOnly) have a similar percentage of dynamic units as the C2D network, suggesting that these models do not capture sufficiently complex dynamic representations. While the static and dynamic biases in 3D CNNs do not fluctuate significantly over the different layers, the transformer-based architectures encode an increasing amount of dynamic information up until about halfway through the model, at which point the pattern tapers off and even reverses slightly.

For the SlowFast network, we measure the biases on the representations for the slow and the fast branches  before fusion of the features. As shown in Fig.~\ref{fig:stagewise_ar_all_models} (b), the fast branch has a nontrivial number of dynamic units. A key component of the SlowFast network is the fusion branches that aim to transfer information from the fast branch to the slow branch. This transfer is accomplished by concatenating the slow and fast features followed by a time-strided convolution. Comparing the dynamic and static between the SlowOnly and SlowFast (slow) branch can reveal whether dynamic information is transferred between the pathways. The addition of the fast branch increases the dynamic units in the slow pathway by 3.3\% as early as stage two, showing the ability of the two-stream architecture to capture dynamics.



\noindent \textbf{Unit-wise analysis.} 
We now examine individual units using our unit-wise metric, (Eq.~\ref{eq:ind_bias_scores_diff_b}),
with $\lambda=0.5$ and report the results in Fig.~\ref{fig:stagewise_ar_all_models} (right). Considering the final representation before the fully connected layer (\ie stage 5), all single stream models, other than FastOnly, contain mainly \textit{static} and \textit{joint} units. In contrast, the FastOnly model and SlowFast-Fast branch produce a nontrivial number of \textit{dynamic} units. Another finding consistent with the results from Fig.~\ref{fig:stagewise_ar_all_models} (left), is that the Fast model extracts more dynamic information \textit{when trained jointly with the Slow branch} than when trained independently. Studying the earlier layers (Fig.~\ref{fig:stagewise_ar_all_models} (right)) confirms the previous findings that the SlowFast model's cross connections are successful at transferring dynamic information from the fast branch to the slow branch throughout the network. SF-Slow (\ie the slow branch) produces a small but notable number of dynamic units early in the network. Interestingly, SF-Fast (\ie the fast branch) gradually produces more dynamic units as the representation flows deeper through the network. All single stream models other than the FastOnly model produces mainly static and joint units in all layers. The MViT and TimeSformer produce mainly static and joint units; however, the ratio of these units changes through the layers of the MViT model while it remains stable through the layers of the TimeSformer.

\vspace{-0.2cm}
\subsubsection{Automatic video object segmentation} 
\label{sec:vos_archs}
\noindent\textbf{Architectures.} We study the dynamic and static biases of two-stream fusion AVOS models that take optical flow and an RGB image as input, with different types of cross connections: (i) FusionSeg~\cite{jain2017fusionseg} with no cross connections, (ii) MATNet~\cite{zhou2020motion} with motion-to-appearance cross connections and (iii) RTNet~\cite{ren2021reciprocal} with bidirectional cross connections. We concentrate on two-stream models because they currently are the strongest AVOS performers. For both MATNet~\cite{zhou2020motion} and RTNet~\cite{ren2021reciprocal}, we use the models provided by the authors without further fine-tuning. 
We use a stylized version of DAVIS16 in our analysis to evaluate the static and dynamic biases for the previous models, with stylization according to Sec.~\ref{sec:sampling}. In the case of both motion and appearance streams, we analyse features after cross connections, if present. Similarly, we analyze the features extracted after the fusion layers in all models, if present. 



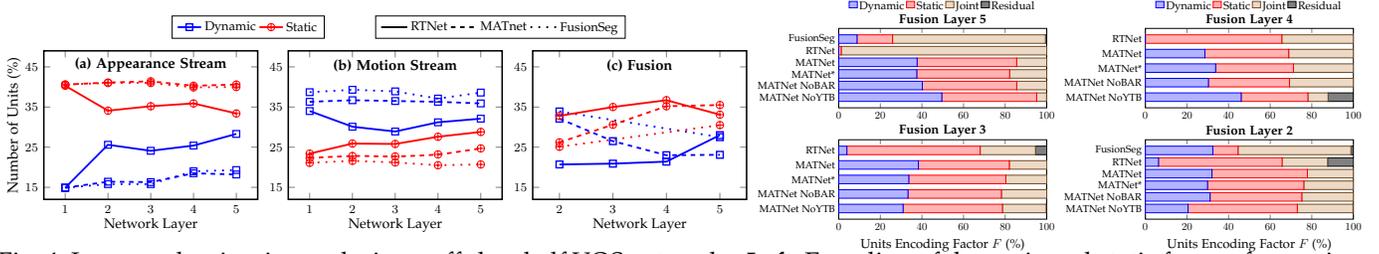
\begin{figure*} [t]
	\begin{center}
     \centering 
     \begin{minipage}{0.55\textwidth}
		\resizebox{\textwidth}{!}{
     
     \begin{tikzpicture} \ref{legend_color_vos} \ref{legend_shades_vos}
     \begin{groupplot}[group style = {group size = 3 by 1, horizontal sep = 20pt}, width = 5.0cm, height = 5.0cm]
     \nextgroupplot[
                 line width=1.0,
                 title={\textbf{(a) Appearance Stream}},
                 title style={at={(axis description cs:0.5,0.93)},anchor=north,font=\normalsize},
                 xlabel={Network Layer},
                 ylabel={Number of Units (\%)},
                 xmin=0.5, xmax=5.5,
                 ymin=12, ymax=49,
                 xtick={1,2,3,4,5},
                 ytick={15,25,35,45},
                 x tick label style={font=\footnotesize},
                 y tick label style={font=\footnotesize},
                 x label style={at={(axis description cs:0.5,0.07)},anchor=north,font=\small},
                 y label style={at={(axis description cs:0.17,.5)},anchor=south,font=\normalsize},
                 width=6.5cm,
                 height=5cm,
                 ymajorgrids=false,
                 xmajorgrids=false,
                 major grid style={dotted,green!20!black},
                 legend style={
                nodes={scale=0.87, transform shape},
                cells={anchor=west},
                legend style={at={(4.7,3.7)},anchor=south,row sep=0.01pt}, font =\normalsize},
                legend image post style={scale=0.9},
                legend columns=2,
                legend to name=legend_color_vos,
             ]
            \addlegendimage{line width=1.2pt,mark=square,mark options={line width=0.9pt,scale=1.2,solid},color=blue}
            \addlegendentry[color=black]{Dynamic}
            \addlegendimage{line width=1.2pt,mark=oplus,color=red,mark options={line width=0.7pt,scale=1.3,solid}}
            \addlegendentry[color=black]{Static}
            
             \addplot[line width=1.2pt, mark options={line width=0.8pt,scale=1.1,solid}, color=blue, mark=square, style=solid]
                     coordinates {(1,14.9)
                                  (2,25.6)
                                  (3,24.1)
                                  (4,25.4)
                                  (5,28.3)};

            \addplot[line width=1.2pt,mark options={line width=0.5pt,scale=1.2,solid}, color=red, mark=oplus, style=solid]
                     coordinates {(1,40.3)
                                  (2,34.1)
                                  (3,35.2)
                                  (4,35.9)
                                  (5,33.4)};
            
            \addplot[line width=1.2pt,mark options={line width=0.8pt,scale=1.1,solid}, color=blue, mark=square, style=dashed]
                     coordinates {(1,14.9)
                                  (2,16.4)
                                  (3,16.3)
                                  (4,18.5)
                                  (5,18.2)};
            \addplot[line width=1.2pt,mark options={line width=0.5pt,scale=1.2,solid}, color=red, mark=oplus, style=dashed]
                     coordinates {(1,40.6)
                                  (2,41.1)
                                  (3,41.4)
                                  (4,40.3)
                                  (5,40.6)};
                                  
            \addplot[line width=1.2pt,mark options={line width=0.8pt,scale=1.1,solid}, color=blue, mark=square, style=loosely dotted]
                     coordinates {(1,14.9)
                                  (2,15.8)
                                  (3,15.8)
                                  (4,19.0)
                                  (5,19.3)};
                                  
             \addplot[line width=1.2pt, mark options={line width=0.5pt,scale=1.2,solid}, color=red, mark=oplus, style=loosely dotted]
                     coordinates {(1,40.6)
                                  (2,41.0)
                                  (3,41.0)
                                  (4,39.9)
                                  (5,40.0)};

                                  

            
            \nextgroupplot[ line width=1.0,
                 title={\textbf{(b) Motion Stream}},
                 title style={at={(axis description cs:0.5,0.92)},anchor=north,font=\normalsize},
                 xlabel={Network Layer},
                 ylabel={},
                 xmin=0.5, xmax=5.5,
                 ymin=12, ymax=49,
                 xtick={1,2,3,4,5},
                ytick={15,25,35,45},
                x tick label style={font=\footnotesize},
                 y tick label style={font=\footnotesize},
                 x label style={at={(axis description cs:0.5,0.07)},anchor=north,font=\small},
                 width=6.5cm,
                 height=5cm,
                 ymajorgrids=false,
                 xmajorgrids=false,
                 major grid style={dotted,green!20!black},
                 legend style={
                     legend style={row sep=0.1pt},
                    nodes={scale=0.87, transform shape},
                    legend columns=-1,
                    cells={anchor=west},
                    legend style={at={(10.5,3.7)},anchor=south,row sep=0.01pt}, font =\normalsize},
                    legend to name=legend_shades_vos,
                 ]
             
            \addlegendimage{line width=1.2pt,color=black, style=solid}
            \addlegendentry[color=black]{RTNet}
            \addlegendimage{line width=1.2pt,color=black, style=dashed}
            \addlegendentry[color=black]{MATnet}
            \addlegendimage{line width=1.2pt,color=black, style=loosely dotted}
            \addlegendentry[color=black]{FusionSeg}

            \addplot[line width=1.2pt, mark options={line width=0.8pt,scale=1.1,solid}, color=blue, style=loosely dotted, mark=square]
                     coordinates {(1,38.7)
                                  (2,39.3)
                                  (3,38.9)
                                  (4,37.1)
                                  (5,38.6)};
                                 
            \addplot[line width=1.2pt, mark options={line width=0.8pt,scale=1.1,solid}, color=blue, style=dashed, mark=square]
                     coordinates {(1,36.3)
                                  (2,36.7)
                                  (3,36.5)
                                  (4,36.3)
                                  (5,35.9)};
            \addplot[line width=1.2pt, mark options={line width=0.8pt,scale=1.1,solid}, color=blue, style=solid, mark=square]
                     coordinates {(1,34.0)
                                  (2,30.1)
                                  (3,28.9)
                                  (4,31.2)
                                  (5,32.1)};
                                  
            \addplot[line width=1.2pt, mark options={line width=0.5pt,scale=1.2,solid}, color=red, style=loosely dotted, mark=oplus]
                     coordinates {(1,21.1)
                                  (2,21.6)
                                  (3,21.2)
                                  (4,20.5)
                                  (5,20.7)};
            \addplot[smooth, line width=1.2pt, mark options={line width=0.5pt,scale=1.2,solid}, color=red, style=dashed, mark=oplus]
                     coordinates {(1,22.3)
                                  (2,22.8)
                                  (3,22.7)
                                  (4,23.2)
                                  (5,24.7)};
            
            \addplot[line width=1.2pt,  mark options={line width=0.5pt,scale=1.2,solid}, color=red, style=solid, mark=oplus]
                     coordinates {(1,23.4)
                                  (2,25.9)
                                  (3,25.8)
                                  (4,27.6)
                                  (5,28.8)};
        
        \nextgroupplot[ line width=1.0,
                 title={\textbf{(c) Fusion}},
                 title style={at={(axis description cs:0.5,0.92)},anchor=north,font=\normalsize},
                 xlabel={Network Layer},
                 ylabel={},
                 xmin=1.5, xmax=5.5,
                 ymin=12, ymax=49,
                 xtick={1,2,3,4,5},
                ytick={15,25,35,45},
                x tick label style={font=\footnotesize},
                 y tick label style={font=\footnotesize},
                 x label style={at={(axis description cs:0.5,0.07)},anchor=north,font=\small},
                 width=6.5cm,
                 height=5cm,
                 ymajorgrids=false,
                 xmajorgrids=false,
                 major grid style={dotted,green!20!black},
             ]
            \addplot[line width=1.2pt, mark options={line width=0.9pt,scale=1.1,solid}, color=blue, style=loosely dotted, mark=square]
                     coordinates {(2,33.9)
                                  (5,27.5)};
                                 
            \addplot[line width=1.2pt, mark options={line width=0.9pt,scale=1.1,solid}, color=blue, style=dashed, mark=square]
                     coordinates {(2,32.1)
                                  (3,26.5)
                                  (4,23.0)
                                  (5,23.1)};
             \addplot[line width=1.2pt, mark options={line width=0.9pt,scale=1.1,solid}, color=blue, mark=square]
                     coordinates {(2,20.7)
                                  (3,20.9)
                                  (4,21.4)
                                  (5,28.0)};
                                  
            \addplot[line width=1.2pt,  mark options={line width=0.5pt,scale=1.2,solid}, color=red, style=loosely dotted, mark=oplus]
                     coordinates {(2,25.1)
                                  (5,30.5)};
            \addplot[line width=1.2pt,   mark options={line width=0.5pt,scale=1.2,solid}, color=red, style=dashed, mark=oplus]
                     coordinates {(2,26.2)
                                  (3,30.6)
                                  (4,35.2)
                                  (5,35.5)};
            
            \addplot[line width=1.2pt,  mark options={line width=0.5pt,scale=1.2,solid}, color=red, mark=oplus]
                     coordinates {(2,32.8)
                                  (3,35.0)
                                  (4,36.7)
                                  (5,33.1)};

\end{groupplot}
\vspace{-0.5cm}
        
\end{tikzpicture}
}
\end{minipage}%
\begin{minipage}{0.45\textwidth}
\resizebox{1.0\textwidth}{!}{


\begin{tikzpicture}
\begin{axis} [xbar stacked,
    width=\axisdefaultwidth,
    height=4.2cm,
    bar width = 8pt,
    xmin = 0,
    xmax = 100,
    title = \textbf{Fusion Layer 5},
    title style={at={(axis description cs:0.5,1.15)},anchor=north,font=\large},
    ytick=data,
    legend style={
         draw=none,
         legend style={row sep=0.1pt},
        nodes={scale=0.87, transform shape},
        legend columns=-1,
        cells={anchor=west},
        legend style={at={(0.5,1.4)},anchor=north,row sep=0.01pt}, font =\large},
    symbolic y coords={MATNet NoYTB, MATNet NoBAR, MATNet*, MATNet, RTNet, FusionSeg},
    enlarge x limits = {value = .1},
    enlarge y limits={abs=10pt}
]

\addplot coordinates {(8.7890625,FusionSeg) (37.6953125,MATNet) (37.5244140625,MATNet*) (40.185546875,MATNet NoBAR) (49.560546875,MATNet NoYTB) (0.0,RTNet)};
\addplot coordinates {(17.08984375,FusionSeg) (47.8515625,MATNet)  (44.62890625,MATNet*) (45.3857421875,MATNet NoBAR) (45.60546875,MATNet NoYTB) (1.171875,RTNet)};
\addplot coordinates {(73.486328125,FusionSeg) (14.453125,MATNet) (17.8466796875,MATNet*) (14.4287109375,MATNet NoBAR) (4.736328125,MATNet NoYTB) (98.828125,RTNet)};
\addplot coordinates {(0.634765625,FusionSeg) (0.0,MATNet) (0.0,MATNet*) (0.0,MATNet NoBAR) (0.09,MATNet NoYTB) (0.0,RTNet)};

\legend {Dynamic,Static,Joint,Residual};

\end{axis}
\end{tikzpicture} 
\hfill
\begin{tikzpicture}
\begin{axis} [xbar stacked,
    width=\axisdefaultwidth,
    height=4.2cm,
    bar width = 8pt,
    xmin = 0,
    xmax = 100,
    title = \textbf{Fusion Layer 4},
    title style={at={(axis description cs:0.5,1.15)},anchor=north,font=\large},
    ytick=data,
    legend style={
         draw=none,
         legend style={row sep=0.1pt},
        nodes={scale=0.87, transform shape},
        legend columns=-1,
        cells={anchor=west},
        legend style={at={(0.5,1.4)},anchor=north,row sep=0.01pt}, font =\large},
    symbolic y coords={MATNet NoYTB, MATNet NoBAR, MATNet*, MATNet, RTNet, FusionSeg},
    enlarge x limits = {value = .1},
    enlarge y limits={abs=10pt}
]

\addplot coordinates {(28.6,MATNet) (33.7890625,MATNet*) (30.3,MATNet NoBAR) (46.044921875,MATNet NoYTB) (0.0,RTNet)};
\addplot coordinates {(40.3,MATNet) (37.40234375,MATNet*) (39.0,MATNet NoBAR) (32.080078125,MATNet NoYTB) (65.6,RTNet)};
\addplot coordinates {(31.1,MATNet) (28.80859375,MATNet*) (30.7,MATNet NoBAR) (9.716796875,MATNet NoYTB) (34.4,RTNet)};
\addplot coordinates {(0.0,MATNet) (0.0,MATNet*) (0.0,MATNet NoBAR) (12.158203125,MATNet NoYTB) (0.0,RTNet)};

\legend {Dynamic,Static,Joint,Residual};

\end{axis}
\end{tikzpicture}
}
\vfill
\resizebox{1.0\textwidth}{!}{
\begin{tikzpicture}
\begin{axis} [xbar stacked,
    width=\axisdefaultwidth,
    height=4.2cm,
    bar width = 8pt,
    xmin = 0,
    xmax = 100,
    title = \textbf{Fusion Layer 3},
    title style={at={(axis description cs:0.5,1.15)},anchor=north,font=\large},
    xlabel = Units Encoding Factor $F$ (\%),
    xlabel style = {font=\large},
    ytick=data,
    legend style={
			area legend,
			at={(0.5,1)},
			anchor=north,
			legend columns=-1},
    symbolic y coords={MATNet NoYTB, MATNet NoBAR, MATNet*, MATNet, RTNet, FusionSeg},
    enlarge x limits = {value = .1},
    enlarge y limits={abs=10pt}
]
\addplot coordinates {(38.3,MATNet) (33.69140625,MATNet*) (33.3,MATNet NoBAR) (30.95703125,MATNet NoYTB) (3.9,RTNet)};
\addplot coordinates {(43.7,MATNet) (46.6796875,MATNet*) (44.9,MATNet NoBAR) (47.75390625,MATNet NoYTB) (64.1,RTNet)};
\addplot coordinates {(18.1,MATNet) (19.62890625,MATNet*) (21.8,MATNet NoBAR) (20.99609375,MATNet NoYTB) (26.6,RTNet)};
\addplot coordinates {(0.0,MATNet) (0.0,MATNet*) (0.0,MATNet NoBAR) (0.29296875,MATNet NoYTB) (5.5,RTNet)};


\end{axis}
\end{tikzpicture}
\begin{tikzpicture}
\begin{axis} [xbar stacked,
     width=\axisdefaultwidth,
    height=4.2cm,
    bar width = 8pt,
    xmin = 0,
    xmax = 100,
    title = \textbf{Fusion Layer 2},
    title style={at={(axis description cs:0.5,1.15)},anchor=north,font=\large},
    xlabel = Units Encoding Factor $F$ (\%),
    xlabel style = {font=\large},
    ytick=data,
    legend style={
			area legend,
			at={(0.5,1)},
			anchor=north,
			legend columns=-1},
    symbolic y coords={MATNet NoYTB, MATNet NoBAR, MATNet*, MATNet, RTNet, FusionSeg},
    enlarge x limits = {value = .1},
    enlarge y limits={abs=10pt}
]
\addplot coordinates {(32.0,MATNet) (29.8828125,MATNet*) (31.1,MATNet NoBAR) (20.5078125,MATNet NoYTB) (6.3,RTNet) (32.421875,FusionSeg)};
\addplot coordinates {(45.9,MATNet) (46.2890625,MATNet*) (44.1,MATNet NoBAR) (52.5390625,MATNet NoYTB) (59.4,RTNet) (12.109375,FusionSeg)};
\addplot coordinates {(22.1,MATNet) (23.828125,MATNet*) (24.8,MATNet NoBAR) (26.7578125,MATNet NoYTB) (21.9,RTNet) (54.296875,FusionSeg)};
\addplot coordinates {(0.0,MATNet) (0.0,MATNet*) (0.0,MATNet NoBAR) (0.1953125,MATNet NoYTB) (12.5,RTNet) (1.171875,FusionSeg)};


\end{axis}
\end{tikzpicture}
}
\end{minipage}
	\end{center}
	\vspace{-0.6cm}
	\caption{Layer and unit-wise analysis on off-the-shelf VOS networks. \textbf{Left}: Encoding of dynamic and static factors for motion, appearance streams and fusion layers in FusionSeg~\cite{jain2017fusionseg}, MATNet~\cite{zhou2020motion} and RTNet~\cite{ren2021reciprocal} using the layer-wise metric, (Eq.~\ref{eq:biasscores}). Fusion layers are mostly biased toward the static factor. \textbf{Right}: Individual units analysis for the three models for fusion layer 5 using the unit-wise metric, (Eq.~\ref{eq:ind_bias_scores_diff_b}). MATNet has the largest number of dynamic units.}\label{fig:stagewise_vos}
	\vspace{-0.64cm}
\end{figure*}

\noindent\textbf{Implementation details.} For a fair comparison with MATNet and RTNet, that fuse motion and appearance features in the intermediate representations, we use a modified version of FusionSeg~\cite{jain2017fusionseg} trained on DAVIS16~\cite{Perazzi2016} in our analysis. Our modified model follows an encoder-decoder approach~\cite{chen2018encoder}, resulting in two fusion layers at stages two and five. Our FusionSeg model is trained with a batch size of eight, using SGD with learning rate 0.001, a momentum of 0.9, a ``poly'' learning rate policy using a power of 0.9 and weight decay $1\times10^{-4}$. As we want to isolate the effect of training datasets, we do not perform pretraining with the motion stream, as proposed in the original paper~\cite{jain2017fusionseg}.
The MATNet variants are trained with two GPUs in parallel with batch size six (the original MATNet used a batch size of two and a single GPU). For the rest of the hyperparameters and training procedure, we follow the original work~\cite{zhou2020motion}. We denote the original model provided by the authors as ``MATNet'', while our reproduction of MATNet as ``MATNet*''. We analyze MATNet trained only on DAVIS16 (\ie without any Youtube-VOS data), which we call ``MATNet NoYTB'' and also MATNet trained without its boundary-aware refinement module and boundary loss, denoted as ``MATNet NoBAR''. 

\noindent \textbf{Layer-wise analysis.} Figure~\ref{fig:stagewise_vos} (left), shows the layer-wise analysis for the motion and appearance streams as well as the fusion layers according to our layer-wise metric, (Eq.~\ref{eq:biasscores}).
Similar to our finding with the action recognition models in Sec.~\ref{sec:ar_models}, the majority of the video object segmentation models are biased toward the \textit{static} factor in the fusion layers (\ie fusion layers three, four and five). We observe an increase in the dynamic bias in the appearance stream as we go deeper in the network, especially for RTNet. In contrast, the static and dynamic biases in the motion streams of both FusionSeg and MATNet are somewhat consistent throughout layers. Interestingly, in RTNet, the \textit{static} bias increases as the representation goes deeper in the network. This result likely stems from the bidirectional cross-connections in RTNet.

\noindent \textbf{Unit-wise analysis.} 
The individual unit analysis for these models obtained using our unit-wise metric, (Eq.~\ref{eq:ind_bias_scores_diff_b}), with $\lambda=0.5$
is shown in Fig.~\ref{fig:stagewise_vos} (right). Looking at the final representation of the models (\ie fusion layer 5), we observe that MATNet has more dynamic and static units compared to RTNet and FusionSeg, which both show a greater number of jointly encoding units. This pattern indicates that cross connections, as present in MATNet, can lead to an increase in the specialized units that encode the static and dynamic factors in the late fusion layers (here we define `specialized' as simply a unit's dedication to encoding a specific factor). The ``MATNet NoBAR'' and ``MATNet NoYTB'' results are consistent, and confirm that the source behind such an increase is not the BAR module or YTB training. The earlier fusion layers in Fig.~\ref{fig:stagewise_vos} (right) also show that MATNet captures the most dynamic units. In fusion layer two, FusionSeg appears on par with MATNet in terms of dynamic units, but has fewer static units and more joint units. In comparison, RTNet tends to have the most unbalanced units of all three models, which become skewed toward joint encoding units in the late fusion layers. 
These patterns show that over all fusion layers, MATNet generally has a more balanced ratio of dynamic and static units and more dynamic units than other models.
It also shows that MATNet trained on solely DAVIS'16 exhibits a similar pattern of capturing more dynamic units than other models, but has less dynamic units in the earlier layers, unlike the original MATNet. This result suggests models with cross connections that are not pretrained on saliency segmentation datasets are beneficial to the model's encoding capabilities when the goal is to capture dynamics, \eg camouflage object segmentation.

\begin{figure} [t]
	\begin{center}
     \centering 
     \begin{minipage}{0.185\textwidth}
		\resizebox{\textwidth}{!}{
     \begin{tikzpicture}  \ref{legend_color_instance_vos}
     \begin{axis} [
     line width=1.0,
                 title={\textbf{VIS - Layerwise}},
                 title style={at={(axis description cs:0.5,0.95)},anchor=north,font=\normalsize},
                 xlabel={Network Layer},
                 ylabel={Number of Units (\%)},
                 xmin=0.5, xmax=12.5,
                 ymin=20, ymax=45,
                 xtick={2,4,6,8,10,12},
                 ytick={15,20,25,30,35,40,45},
                 x tick label style={font=\small},
                 y tick label style={font=\small},
                 x label style={at={(axis description cs:0.5,0.07)},anchor=north,font=\normalsize},
                 y label style={at={(axis description cs:0.17,.5)},anchor=south,font=\normalsize},
                 xtick pos=left,
                 ytick pos=bottom,
                 width=6.5cm,
                 height=5cm,
                 ymajorgrids=false,
                 xmajorgrids=false,
                 major grid style={dotted,green!20!black},
                 legend style={
                     nodes={scale=0.7, transform shape},
                     cells={anchor=west},
                     legend style={at={(3,1)},anchor=south}, font =\small},
                     legend columns=2,
                     legend entries={[black]R50-Dyn,[black]R50-Stat,[black]R101-Dyn,[black]R101-Stat},
                    legend to name=legend_color_instance_vos,
        ]

            \addplot[line width=1.2pt,mark options={line width=0.8pt,scale=1,solid}, color=blue, mark=square, style=dashdotted]
                     coordinates {(1,23.7)
                                  (2,24.2)
                                  (3,24.2)
                                  (4,25.0)
                                  (5,25.0)
                                  (6,24.7)
                                  (7,22.9)
                                  (8,23.4)
                                  (9,22.9)
                                  (10,23.2)
                                  (11,23.4)
                                  (12,22.9)};
                                  
            \addplot[line width=1pt, mark options={line width=0.5pt,scale=1.2,solid}, color=red, mark=oplus, style=dashdotted]
                     coordinates {(1,37.5)
                                  (2,36.5)
                                  (3,36.5)
                                  (4,35.9)
                                  (5,35.9)
                                  (6,36.5)
                                  (7,38.8)
                                  (8,38.3)
                                  (9,38.3)
                                  (10,38.3)
                                  (11,38.0)
                                  (12,38.5)};

            \addplot[line width=1pt,mark options={line width=0.8pt,scale=1.1,solid}, color=blue, mark=triangle, style=solid]
                     coordinates {(1,25.3)
                                  (2,23.4)
                                  (3,23.4)
                                  (4,23.2)
                                  (5,23.2)
                                  (6,24.0)
                                  (7,23.2)
                                  (8,22.4)
                                  (9,22.4)
                                  (10,22.4)
                                  (11,22.4)
                                  (12,22.7)};
                                  
            \addplot[line width=1pt, mark options={line width=0.5pt,scale=1.2,solid}, color=red, mark=+, style=solid]
                     coordinates {(1,36.2)
                                  (2,37.2)
                                  (3,37.0)
                                  (4,37.2)
                                  (5,36.7)
                                  (6,36.5)
                                  (7,38.0)
                                  (8,38.5)
                                  (9,38.5)
                                  (10,38.5)
                                  (11,38.8)
                                  (12,38.3)};
                                  
        \end{axis}
\end{tikzpicture}
}
\end{minipage}%
\begin{minipage}{0.31\textwidth}

\resizebox{1.0\textwidth}{!}{

\begin{tikzpicture}
\begin{axis} [xbar stacked,
    width=8cm,
    height=2.6cm,
    bar width = 8pt,
    xmin = 0,
    xmax = 100,
    title = \textbf{Layer 3},
    title style={at={(axis description cs:0.5,1.5)},anchor=north,font=\LARGE},
    x tick label style={font=\large},
    y tick label style={font=\large},
    x label style={at={(axis description cs:0.5,0.07)},anchor=north,font=\large},
    y label style={at={(axis description cs:0.17,.5)},anchor=south,font=\large},
    ytick=data,
    legend style={
         draw=none,
         legend style={row sep=0.1pt},
        nodes={scale=0.87, transform shape},
        legend columns=-1,
        cells={anchor=west},
        legend style={at={(0.5,2.3)},anchor=north,row sep=0.01pt}, font =\Large},
    symbolic y coords={R101,R50},
    enlarge x limits = {value = .1},
    enlarge y limits={abs=8pt}
]

\addplot coordinates {(0,R101) (0,R50)};
\addplot coordinates {(62.8,R101) (64.1,R50)};
\addplot coordinates {(37.2,R101) (35.9,R50)};
\addplot coordinates {(0.0,R101) (0.0,R50)};

\legend {Dynamic,Static,Joint,Residual};

\end{axis}
\end{tikzpicture} 
\hfill
\begin{tikzpicture}
\begin{axis} [xbar stacked,
    width=8cm,
    height=2.6cm,
    bar width = 8pt,
    xmin = 0,
    xmax = 100,
    title = \textbf{Layer 6},
    title style={at={(axis description cs:0.5,1.5)},anchor=north,font=\LARGE},
    x tick label style={font=\large},
    y tick label style={font=\large},
    x label style={at={(axis description cs:0.5,0.07)},anchor=north,font=\large},
    y label style={at={(axis description cs:0.17,.5)},anchor=south,font=\large},
    ytick=data,
    legend style={
         draw=none,
         legend style={row sep=0.1pt},
        nodes={scale=0.87, transform shape},
        legend columns=-1,
        cells={anchor=west},
        legend style={at={(0.5,2.3)},anchor=north,row sep=0.01pt}, font =\Large},
    symbolic y coords={R101,R50},
    enlarge x limits = {value = .1},
    enlarge y limits={abs=8pt}
]

\addplot coordinates {(0,R101) (0,R50)};
\addplot coordinates {(63.0,R101) (49.2,R50)};
\addplot coordinates {(37.0,R101) (50.8,R50)};
\addplot coordinates {(0.0,R101) (0.0,R50)};

\legend {Dynamic,Static,Joint,Residual};

\end{axis}
\end{tikzpicture}
}
\vfill
\resizebox{1.0\textwidth}{!}{
\begin{tikzpicture}
\begin{axis} [xbar stacked,
    width=8cm,
    height=2.6cm,
    bar width = 8pt,
    xmin = 0,
    xmax = 100,
    title = \textbf{Layer 9},
    title style={at={(axis description cs:0.5,1.5)},anchor=north,font=\LARGE},
    x tick label style={font=\large},
    y tick label style={font=\large},
    x label style={at={(axis description cs:0.5,0.0)},anchor=north,font=\Large},
    xlabel = Units Encoding Factor $F$ (\%),
    ytick=data,
    legend style={
			area legend,
			at={(0.5,1)},
			anchor=north,
			legend columns=-1},
    symbolic y coords={R101,R50},
    enlarge x limits = {value = .1},
    enlarge y limits={abs=8pt}
]
\addplot coordinates {(0,R101) (0,R50)};
\addplot coordinates {(64.3,R101) (53.4,R50)};
\addplot coordinates {(35.7,R101) (46.6,R50)};
\addplot coordinates {(0.0,R101) (0.0,R50)};


\end{axis}
\end{tikzpicture}
\begin{tikzpicture}
\begin{axis} [xbar stacked,
    width=8cm,
    height=2.6cm,
    bar width = 8pt,
    xmin = 0,
    xmax = 100,
    title = \textbf{Layer 12},
    title style={at={(axis description cs:0.5,1.5)},anchor=north,font=\LARGE},
    xlabel = Units Encoding Factor $F$ (\%),
    x tick label style={font=\large},
    y tick label style={font=\large},
    xlabel style={at={(axis description cs:0.5,0.0)},anchor=north,font=\Large},
    ytick=data,
    legend style={
			area legend,
			at={(0.5,1)},
			anchor=north,
			legend columns=-1},
    symbolic y coords={R101,R50},
    enlarge x limits = {value = .1},
    enlarge y limits={abs=8pt}
]
\addplot coordinates {(0,R101) (0,R50)};
\addplot coordinates {(62.0,R101) (50.5,R50)};
\addplot coordinates {(38.0,R101) (49.5,R50)};
\addplot coordinates {(0.0,R101) (0.0,R50)};


\end{axis}
\end{tikzpicture}
}
\end{minipage}
	\end{center}
	\vspace{-0.6cm}
	\caption{Layer and unit-wise analysis on off-the-shelf state-of-the-art VIS models. \textbf{Left}: Encoding of dynamic and static factors for motion, appearance streams and fusion layers in VisTR-R50 and VisTR-101~\cite{wang2021end} using the layer-wise metric, (Eq.~\ref{eq:biasscores}). All layers are biased toward the static factor. \textbf{Right}: Individual units analysis for the two VisTR variants using the unit-wise metric, (Eq.~\ref{eq:ind_bias_scores_diff_b}).}\label{fig:inst_vos_arch}
	\vspace{-0.3cm}
\end{figure}
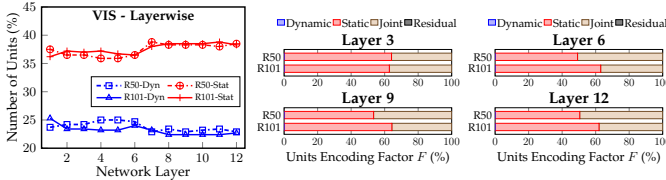

\vspace{-0.3cm}
\subsubsection{Video instance segmentation} 
\label{sec:vis_archs}

\noindent\textbf{Architectures.} For the VIS task, we conduct our analysis on two versions of the state-of-the-art VisTR~\cite{wang2021end} model. VisTR is an end-to-end trained encoder-decoder transformer based architecture with a 2D CNN backbone used for the initial feature extraction. Both the encoder and decoder have six layers for which we compute the static and dynamic biases.

\noindent\textbf{Implementation details.} All pre-trained models are retrieved from the original repository~\cite{wang2021end} with no training done on our end. All VIS models are trained on the YouTube-VIS~\cite{xu2018youtube} dataset. We use the same 2D CNN backbones as per the original paper~\cite{wang2021end}, either a ResNet50 or ResNet101. We perform the static and dynamic estimation using the stylized DAVIS16 dataset.

\noindent\textbf{Layer-wise analysis.} Figure~\ref{fig:inst_vos_arch} (left) shows the results of our layer-wise analysis on two variants of VisTR~\cite{wang2021end}. It is observed that both models are strongly biased toward the static factor. This bias toward static information increases later in the network, \eg at the first transformer layer the ResNet101 variant has static and dynamic biases of 36.2\% and 25.3\%, respectively, and at the last layer has biases of 38.3\% and 22.7\%, respectively. This result likely stems due to the nature of the VIS task that requires tracking and object-level matching across frames.

\noindent\textbf{Unit-wise analysis.} Figure~\ref{fig:inst_vos_arch} (right) shows the results of our layer-wise and per-unit analysis on two variants of VisTR~\cite{wang2021end}. Notably, VisTR with either backbone produces solely static and joint units, however, the ratio of these units differ at various layers of the models. The ResNet50 backbone contains about twice the number of static units compared with joint units for the first three layers, and then converges to an even ratio thereafter. Meanwhile, the ResNet101 backbone contains a larger number of static units throughout the entire architecture. This suggests that the 2D backbone plays a role in the type of information captured for spatiotemporal models and that deeper backbones may capture more static information.

\vspace{-0.3cm}
\subsubsection{Summary and shared insights}
We showed that most of the examined state-of-the-art models for all tasks are biased toward encoding static information.
We also demonstrated the efficacy of two-stream models with motion-to-appearance~\cite{zhou2020motion} (fast-to-slow~\cite{feichtenhofer2019slowfast}) cross connections to enable greater encoding of dynamic information. For the VIS task, we observed that solely static and joint units are produced in the architectures analyzed, however the 2D backbone can influence the ratio between these units. Finally, we documented that the final layers of dynamic biased models are capable of producing more specialized dynamic units compared to the joint units produced by static biased models.

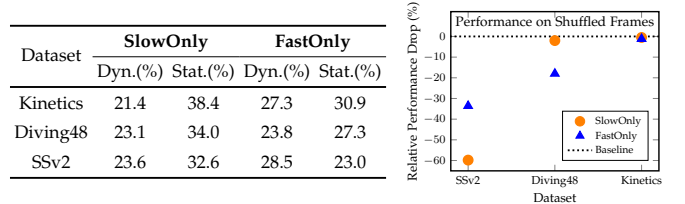
\begin{figure} [t]
 \def\arraystretch{1.35}
 \setlength\tabcolsep{2.4pt}
\centering
\begin{minipage}{0.57\linewidth}
\resizebox{1.0\textwidth}{!}{
	\begin{tabu}{c cccc}
	\tabucline[1pt]{-}
	 \multirow{2}{*}{Dataset}&  \multicolumn{2}{c}{\textbf{SlowOnly}} & \multicolumn{2}{c}{\textbf{FastOnly}} \\
	 \cline{2-3} \cline{4-5}
	  & Dyn.(\%) & Stat.(\%)& Dyn.(\%) & Stat.(\%)\\
	   \tabucline[1pt]{-}
	   Kinetics& 21.4 & 38.4 & 27.3 & 30.9 \\
	   Diving48& 23.1 & 34.0 & 23.8 & 27.3\\
	   SSv2& 23.6 & 32.6 & 28.5 & 23.0 \\
	\tabucline[1pt]{-}
	\end{tabu}}
		\end{minipage} \hfill   
	\begin{minipage}{0.4\linewidth}
	\resizebox{1.0\textwidth}{!}{
	\begin{tikzpicture} \ref{target_legend}
    \begin{axis}[
       line width=1.0,
        title={Performance on Shuffled Frames},
        title style={at={(axis description cs:0.5,0.95)},anchor=north,font=\normalsize},
        xlabel={Dataset},
        ylabel={Relative Performance Drop (\%)},
        ymin=-65, ymax=12,
        ytick={0,-10,-20,-30,-40,-50,-60},
        symbolic x coords={SSv2, Diving48, Kinetics},
        xtick=data,
        x tick label style={font=\footnotesize},
        y tick label style={font=\footnotesize},
        x label style={at={(axis description cs:0.5,0.03)},anchor=north,font=\small},
        y label style={at={(axis description cs:0.12,.5)},anchor=south,font=\small},
        width=6.7cm,
        height=5.5cm,        
        ymajorgrids=false,
        xmajorgrids=false,
        major grid style={dotted,green!20!black},
        legend style={
         nodes={scale=0.9, transform shape},
         cells={anchor=west},
         legend style={at={(3.8,0.25)},anchor=south}, font =\footnotesize},
         legend entries={[black]SlowOnly,[black]FastOnly,[black]Baseline},
        legend to name=target_legend,
    ]
    
    \addplot[only marks,mark size=3.3pt,color=orange,mark=*,]
    coordinates {(SSv2,-59.8) (Diving48,-2.0) (Kinetics,-0.5)};

        
    \addplot[only marks,mark size=3.3pt,color=blue,mark=triangle*,]
        coordinates {(SSv2,-33.6) (Diving48,-18.0) (Kinetics,-1.2)};

    \addplot[line width=1.3pt,black,dotted,sharp plot,update limits=false] 
	    coordinates {([normalized]-10,0)([normalized]10,0)};
    \end{axis}
\end{tikzpicture}}
\end{minipage} 
\vspace{-0.3cm}
\caption{Analyses of biases of action recognition datasets. \textbf{Left:} \textit{Dynamic} and \textit{static} biases using the layer-wise metric, (Eq.~\ref{eq:biasscores}), for models trained on Kinetics-400~\cite{carreira2017quo}, Diving48~\cite{li2018resound} and SSv2~\cite{goyal2017something}. \textbf{Right:} Relative drop in Top 1 Acc (\%) for the SlowOnly and FastOnly models trained with shuffled frames with respect to standard training (\ie baseline).}\label{fig:dataset_compare}
\vspace{-0.7cm}
\end{figure}

\vspace{-0.5cm}
\subsection{Training datasets}\label{sec:dataset_effect}
\vspace{-0.1cm}

\subsubsection{Action recognition}\label{sec:ar_dataset}
\noindent \textbf{Datasets.} With the knowledge that action recognition models often use static context biases in the data to make predictions~\cite{derpanis2012action,choi2019can,ilic2022appearance}, we 
consider datasets in the following evaluations which were designed with the goal of benchmarking a model's ability to capture dynamic information. Two popular datasets of this type are Something-Something-v2~\cite{goyal2017something} (SSv2) and Diving48~\cite{li2018resound}. SSv2 is a fine-grained ego-centric dataset with 174 classes and over 30,000 unique objects. Notably, different actions in SSv2 include similar appearance but different motions, \eg the classes `moving something from right-to-left' and `moving something from left-to-right'. Diving48~\cite{li2018resound} was created to be ``a dataset with no significant biases toward static or short-term motion representations, so that the capability of models to capture long-term dynamics information could be evaluated''~\cite{d48_web}. All actions are a particular type of dive and differ by only a single rotation or flip. We compare Kinetics-400, Diving48 and SSv2 to determine the extent that each dataset requires dynamics for action recognition.

\noindent \textbf{Dataset bias.} We use the layer-wise metric, (Eq.~\ref{eq:biasscores}), to estimate the static and dynamic units captured in the last layer of two models trained on the three datasets, as shown in the table of Fig.~\ref{fig:dataset_compare} (left). We generate Stylized SSv2 and Stylized Diving48 to produce the static and dynamic estimates (and continue using Stylized ActivityNet for Kinetics-400 trained models). We measure the last layer, as the final prediction is made directly from it and thus is most representative of what information the model uses for the final prediction. The SlowOnly and FastOnly architectures follow a similar pattern to that found in Sec.~\ref{sec:architectures}, with the FastOnly consistently capturing more dynamic information. Surprisingly, models trained on Diving48 capture a similar amount of dynamics compared to Kinetics. These results may seem curious at first, as it seems unlikely that models could perform well on Diving48 without dynamic information.

To further understand this result, we conduct a simple experiment, where the model only has static information to learn from. As discussed in Sec.~\ref{sec:sampling}, frame-shuffled videos will have the same static information as a non-shuffled input, but the temporal correlations, and hence dynamic information, will be corrupted. This forces the model to focus on static information for classification. We compare the Top-1 validation accuracy of models trained and validated on shuffled frames to that of models with standard training. Figure~\ref{fig:dataset_compare} (right) shows the results of the SlowOnly and FastOnly networks on Diving48, SSv2 and Kinetics-400, in terms of the relative performance on shuffled frames compared to unshuffled. For a fair comparison, we initialize all models from Kinetics-400. Both models show strong relative performance when trained on shuffled videos for Diving48 and Kinetics-400; however, for SSv2 the classification performance is decreased to a greater extent when trained on shuffled frames. These results show that SSv2 is a better alternative for temporally benchmarking networks.

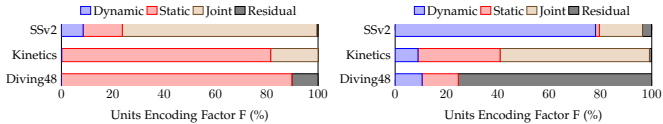
\begin{figure}[t]
    \centering
    \resizebox{0.49\textwidth}{!}{
\begin{tikzpicture}
\begin{axis} [
    width=\axisdefaultwidth,
    height=3.2cm,
    xbar stacked,
    bar width = 10pt,
    xmin = 0,
    xmax = 100,
    xlabel = Units Encoding Factor F (\%),
    ytick=data,
    legend style={
            draw = none,
			area legend,
			at={(0.5,1.4)},
			anchor=north,
			legend columns=-1},
    symbolic y coords={Diving48, Kinetics, SSv2},
    enlarge x limits = {value = 0.1},
]


\addplot coordinates { (0,Diving48) (0.09765625,Kinetics) (8.447265625,SSv2)};
\addplot coordinates { (89.79492188,Diving48) (81.4453125,Kinetics) (15.28320313,SSv2)};
\addplot coordinates { (0,Diving48) (18.45703125,Kinetics) (75.83007813,SSv2)};
\addplot coordinates { (10.20507813,Diving48) (0,Kinetics) (0.439453125,SSv2)};

\legend {Dynamic, Static, Joint, Residual};

\end{axis}
\end{tikzpicture}

\begin{tikzpicture}
\begin{axis} [
    width=\axisdefaultwidth,
    height=3.2cm,
    xbar stacked,
    bar width = 10pt,
    xmin = 0,
    xmax = 100,
    xlabel = Units Encoding Factor F (\%),
    ytick=data,
    legend style={
            draw = none,
			area legend,
			at={(0.5,1.4)},
			anchor=north,
			legend columns=-1},
    symbolic y coords={Diving48, Kinetics, SSv2},
    enlarge x limits = {value = .1},
]


\addplot  coordinates {(10.5,Diving48) (8.9,Kinetics)  (78.1,SSv2)};
\addplot  coordinates {(14.1,Diving48)  (32.0,Kinetics) (1.5,SSv2)};
\addplot  coordinates {(0,Diving48) (58.2,Kinetics) (16.8,SSv2)};
\addplot  coordinates {(75.4,Diving48) (0.8,Kinetics) (3.5,SSv2)};
\legend {Dynamic, Static, Joint, Residual};
\end{axis}
\end{tikzpicture}
}
\vspace{-0.7cm}
\caption{Estimating the dynamic, static, joint and residual units using the unit-wise metric, (Eq.~\ref{eq:ind_bias_scores_diff_b}), for the SlowOnly (\textbf{left}) and FastOnly (\textbf{right}) models on Kinetics-400~\cite{carreira2017quo}, Diving48~\cite{li2018resound} and SSv2~\cite{goyal2017something}. Dynamic units arise from dynamic-biased models (\eg FastOnly) and residual units from training on Diving48.}
\label{fig:ar_dataset_indiv}
\vspace{-0.6cm}
\end{figure}

Figure~\ref{fig:ar_dataset_indiv} shows the individual units (from the last layer) for two models (one static biased, SlowOnly, and one dynamic biased, FastOnly) on Kinetics-400, Diving48 and SSv2. The SlowOnly model trained on Kinetics-400 contains only static and joint units; however, when trained on Diving48 or SSv2, both residual and dynamic units emerge, demonstrating the impact of the training dataset on producing specialized dynamic units. Unlike the SlowOnly model, the FastOnly model contains many dynamic units trained on any dataset, showing the efficacy of the architecture for producing specialized dynamic units. Interestingly, each dataset is unique in the type of units that emerge. Diving48 produces residual units, suggesting there are other factors at play beyond dynamic and static information. On the other hand, SSv2 produces the most dynamic units for both models. 

\subsubsection{Automatic video object segmentation}\label{sec:vos_datasets}

\noindent\textbf{Datasets.} We study the impact of the following four video segmentation datasets on a model's static and dynamic biases:
DAVIS16~\cite{Perazzi2016}, Weakly Labelled ImageNet VID~\cite{jain2017fusionseg}, YouTube-VOS~\cite{xu2018youtube} and TAO-VOS~\cite{Voigtlaender21WACV}.
DAVIS16~\cite{Perazzi2016} is the most widely used benchmark for automatic VOS, with 50 short-temporal extent sequences of two to four seconds and 3455 manually annotated frames. ImageNet VID~\cite{jain2017fusionseg} contains 3251 weakly labelled videos and was used in previous work to pretrain a model's motion stream~\cite{jain2017fusionseg}; in contrast, we use it as a general training dataset (\ie beyond just for motion streams). YouTube-VOS~\cite{xu2018youtube} is another widely used AVOS dataset with 3471 videos in the training set, which is usually combined with DAVIS dataset following~\cite{zhou2020motion} to end up with 14K training images. We assess the two training datasets separately and evaluate how they affect the static and dynamic biases. Finally, TAO-VOS~\cite{Voigtlaender21WACV} contains 626 relatively long videos (36 seconds on average) that are annotated in a hybrid fashion between manually and weakly labelled frames, resulting in 74,187 frames.
We convert the annotations to exclude instances and instead consider foreground/background annotations only.

\noindent\textbf{Dataset bias.} We train our modified FusionSeg with two fusion layers (layers two and five) on the four datasets. We compute the static and dynamic biases for the training datasets using the layer-wise metric,
(\ref{eq:biasscores}), and report the results in Table~\ref{fig:vos_dataset}. The model trained on TAO-VOS has the least amount of static bias out of all four datasets. However, the datasets do not differ much in their dynamic bias.

\begin{table}
\resizebox{0.45\textwidth}{!}{
\begin{tabu}{c cccc}
\tabucline[1pt]{-}
 \multirow{2}{*}{Dataset}&  \multicolumn{2}{c}{\textbf{Fusion Layer 5}} & \multicolumn{2}{c}{\textbf{Fusion Layer 2}}    \\
\cline{2-5}
  & Dyn.(\%) & Stat.(\%)& Dyn.(\%) & Stat.(\%)\\
\tabucline[1pt]{-}
DAVIS & 27.8 & 30.1 & 34.0 & 25.9 \\
YouTube-VOS & 25.2 & 34.2 & 33.5 & 24.9\\
ImageNetVID & 26.4 & 33.1 & 33.0 & 24.6\\
TAO-VOS & 26.4 & 25.8  & 33.7 & 23.2 \\
\tabucline[1pt]{-}
\end{tabu}}
\caption{Biases of video object segmentation datasets using the layer-wise metric, (Eq.~\ref{eq:biasscores}), for FusionSeg's fusion layers five and two, trained on DAVIS16~\cite{Perazzi2016}, YouTube-VOS~\cite{xu2018youtube}, ImageNetVID~\cite{jain2017fusionseg} and TAO-VOS~\cite{Voigtlaender21WACV}.}
\label{fig:vos_dataset}
\vspace{-0.5cm}
\end{table}

\begin{figure}
    \centering
\begin{minipage}{0.35\textwidth}
\resizebox{\textwidth}{!}{
\begin{tikzpicture}
\begin{axis} [
    width=\axisdefaultwidth,
    height=3.2cm,
    xbar stacked,
    bar width = 10pt,
    xmin = 0,
    xmax = 100,
    title = \textbf{Fusion Layer 5},
    ytick=data,
    legend style={
            draw = none,
			area legend,
			at={(0.5,1.4)},
			anchor=north,
			legend columns=-1},
    symbolic y coords={TAO-VOS, ImageNetVID, YouTube-VOS, DAVIS},
    enlarge x limits = {value = .1},
] 
\addplot coordinates { (8.7890625,DAVIS) (0.9765625,ImageNetVID) (0.0,YouTube-VOS) (49.70703125,TAO-VOS)};
\addplot coordinates { ( 17.08984375,DAVIS) ( 20.654296875,ImageNetVID) (41.6,YouTube-VOS) (18.5546875,TAO-VOS)};
\addplot coordinates { ( 73.486328125,DAVIS) (78.173828125,ImageNetVID) (58.4,YouTube-VOS) (22.021484375,TAO-VOS)};
\addplot coordinates { (0.6,DAVIS) (0.1953125,ImageNetVID) (0.0,YouTube-VOS) (9.716796875,TAO-VOS)};
\end{axis}
\end{tikzpicture}}

\resizebox{\textwidth}{!}{
\begin{tikzpicture}
\begin{axis} [
    width=\axisdefaultwidth,
    height=3.2cm,
    xbar stacked,
    bar width = 10pt,
    xmin = 0,
    xmax = 100,
    title = \textbf{Fusion Layer 2},
    xlabel = Units Encoding Factor $F$ (\%),
    ytick=data,
    legend style={
            draw = none,
			area legend,
			at={(0.5,-0.7)},
			anchor=north,
			legend columns=-1},
    symbolic y coords={TAO-VOS, ImageNetVID, YouTube-VOS, DAVIS},
    enlarge x limits = {value = .1},
] 
\addplot coordinates { (32.421875,DAVIS) (60.15625,ImageNetVID) (48.4,YouTube-VOS) (70.703125,TAO-VOS)};
\addplot coordinates { (12.109375,DAVIS) (8.203125,ImageNetVID) (7.0,YouTube-VOS) (3.515625,TAO-VOS)};
\addplot coordinates { (54.296875,DAVIS) (30.46875,ImageNetVID) (42.6,YouTube-VOS) (23.828125,TAO-VOS)};
\addplot coordinates { (1.171875,DAVIS) (1.171875,ImageNetVID) (2.0,YouTube-VOS) (1.953125,TAO-VOS)};
 
\legend {Dynamic, Static, Joint, Residual};
\end{axis}
\end{tikzpicture}}
\end{minipage}%
\begin{minipage}{0.13\textwidth}
\resizebox{\textwidth}{!}{
\includegraphics[width=\textwidth]{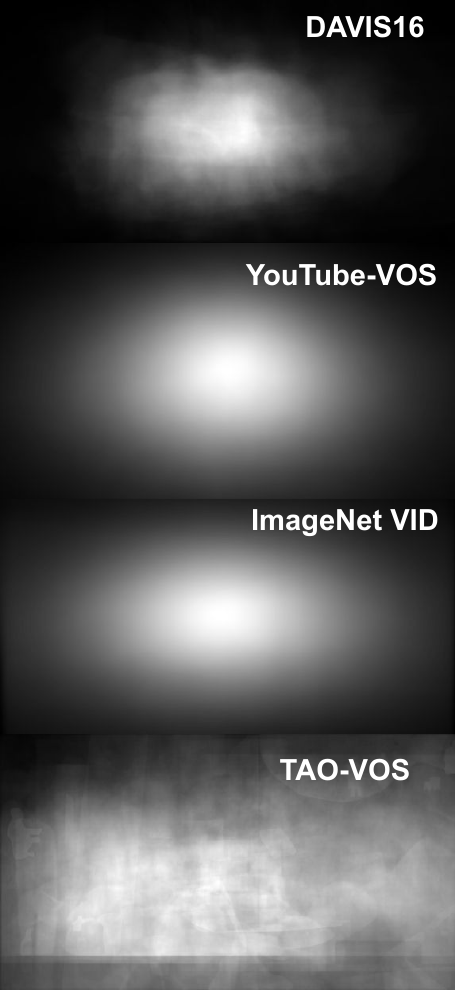}}
\end{minipage}
\vspace{-0.2cm}
\caption{Analyses of biases of AVOS datasets. \textbf{Left}: Estimating the dynamic, static, joint and residual units using the unit-wise metric, (Eq.~\ref{eq:ind_bias_scores_diff_b}), for FusionSeg's fusion layers five and two trained on DAVIS16~\cite{Perazzi2016}, YouTube-VOS~\cite{xu2018youtube}, ImageNetVID~\cite{jain2017fusionseg} and TAO-VOS~\cite{Voigtlaender21WACV}. \textbf{Right}: Center bias plots for the four datasets. The results show the emergence of more dynamic units for both fusion layers when trained on the least center biased dataset (\ie TAO-VOS).}
\label{fig:vos_dataset_indiv}
\vspace{-0.6cm}
\end{figure}
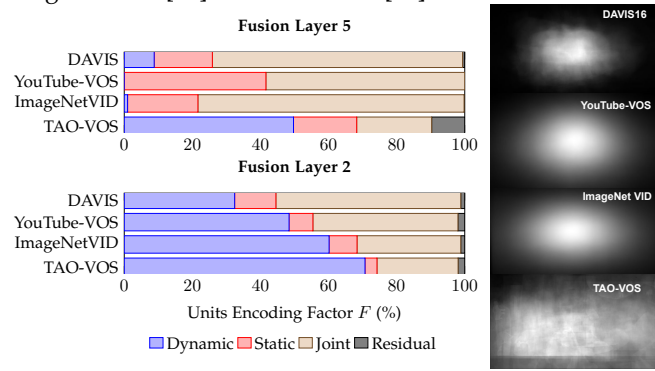

We analyse the datasets in terms of the individual unit analysis using the unit-wise metric, (Eq.~\ref{eq:ind_bias_scores_diff_b}), with $\lambda=0.5$. It is seen in Fig.~\ref{fig:vos_dataset_indiv} (left) that models trained on TAO-VOS produce the highest number of specialized dynamic biased units compared to the other datasets. To explore this matter further, we evaluate the center bias for the four datasets by calculating the average number of groundtruth segmentation masks for each pixel over the entire dataset (normalized to 0-1), with results shown in Fig.~\ref{fig:vos_dataset_indiv} (right). It is seen that for both layers, the percentage of specialized dynamic units is greatest for the dataset that has least center bias, \ie TAO-VOS, as its center bias map is far more diffuse than the others. These results have implications for how the datasets can be used best for different tasks. For example, more general motion segmentation without concern for centering might be better served by training with a dynamic biased dataset (\eg TAO-VOS).

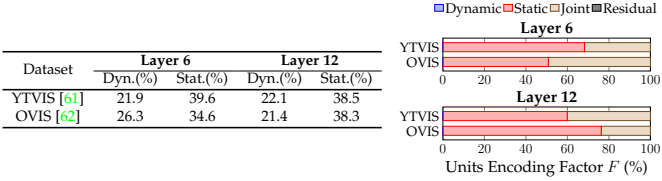
\begin{figure} [t]
	\begin{center}
     \centering 
\begin{minipage}{0.28\textwidth}
\resizebox{1.0\textwidth}{!}{
	\begin{tabu}{c cccc}
	\tabucline[1pt]{-}
	 \multirow{2}{*}{Dataset}&  \multicolumn{2}{c}{\textbf{Layer 6}} & \multicolumn{2}{c}{\textbf{Layer 12}} \\
	 \cline{2-3} \cline{4-5}
	  & Dyn.(\%) & Stat.(\%)& Dyn.(\%) & Stat.(\%)\\
	   \tabucline[1pt]{-}
	   YTVIS~\cite{yang2019video} & 21.9 & 39.6 & 22.1 & 38.5   \\
	   OVIS~\cite{qi2022occluded} & 26.3 & 34.6 & 21.4 & 38.3 \\
	\tabucline[1pt]{-}
	\end{tabu}}
\end{minipage} \hfill   
\begin{minipage}{0.2\textwidth}
\resizebox{1.0\textwidth}{!}{
\begin{tikzpicture}
\begin{axis} [xbar stacked,
    width=8cm,
    height=2.6cm,
    bar width = 8pt,
    xmin = 0,
    xmax = 100,
    title = \textbf{Layer 6},
    title style={at={(axis description cs:0.5,1.4)},anchor=north,font=\Large},
    x tick label style={font=\large},
    y tick label style={font=\large},
    x label style={at={(axis description cs:0.5,0.07)},anchor=north,font=\large},
    y label style={at={(axis description cs:0.17,.5)},anchor=south,font=\large},
    ytick=data,
    legend style={
         draw=none,
         legend style={row sep=0.1pt},
        nodes={scale=0.87, transform shape},
        legend columns=-1,
        cells={anchor=west},
        legend style={at={(0.5,2.2)},anchor=north,row sep=0.01pt}, font =\Large},
    symbolic y coords={OVIS,YTVIS},
    enlarge x limits = {value = .1},
    enlarge y limits={abs=8pt}
]


\addplot coordinates {(0,OVIS) (0,YTVIS)};
\addplot coordinates {(50.8,OVIS) (68.2,YTVIS)};
\addplot coordinates {(49.2,OVIS) (31.8,YTVIS)};
\addplot coordinates {(0.0,OVIS) (0.0,YTVIS)};

\legend {Dynamic,Static,Joint,Residual};

\end{axis}
\end{tikzpicture} 
}
\vfill
\resizebox{1.0\textwidth}{!}{
\begin{tikzpicture}
\begin{axis} [xbar stacked,
    width=8cm,
    height=2.6cm,
    bar width = 8pt,
    xmin = 0,
    xmax = 100,
    title = \textbf{Layer 12},
    title style={at={(axis description cs:0.5,1.4)},anchor=north,font=\Large},
    xlabel = Units Encoding Factor $F$ (\%),
    x tick label style={font=\large},
    y tick label style={font=\large},
    xlabel style={at={(axis description cs:0.5,0.0)},anchor=north,font=\Large},
    ytick=data,
    legend style={
			area legend,
			at={(0.5,1)},
			anchor=north,
			legend columns=-1},
    symbolic y coords={OVIS,YTVIS},
    enlarge x limits = {value = .1},
    enlarge y limits={abs=8pt}
]

\addplot coordinates {(0,OVIS) (0,YTVIS)};
\addplot coordinates {(76.3,OVIS) (59.9,YTVIS)};
\addplot coordinates {(23.7,OVIS) (40.1,YTVIS)};
\addplot coordinates {(0.0,OVIS) (0.0,YTVIS)};


\end{axis}
\end{tikzpicture}
}

\end{minipage}
	\end{center}
	\vspace{-0.3cm}
	\caption{Analyses of biases of VIS datasets. \textbf{Left:} Results for the static and dynamic biases using the layer-wise metric, (Eq.~\ref{eq:biasscores}), for layers 6 and 12 of VisTR-R50 trained on OVIS~\cite{qi2022occluded} and YTVIS~\cite{yang2019video}. \textbf{Right:} Unit-wise analysis, (Eq.~\ref{eq:bias_scores_indv}), of a VisTR-R50 model trained on both datasets for layers 6 and 12.}\label{fig:inst_vos_dataset}
	\vspace{-0.4cm}
\end{figure}


\vspace{-0.2cm}
\subsubsection{Video instance segmentation}\label{sec:vis_dataset}

\noindent \textbf{Datasets.} We study the impact of the following two datasets on a VIS model's static and dynamic biases:  YouTube-VIS 2019 (YTVIS)~\cite{yang2019video} and Occluded Video Instance Segmentation (OVIS)~\cite{qi2022occluded}. YTVIS is a popular VIS benchmark and contains 2,883 videos, 40 categories, 131k total masks, an average of 1.7 instances per video and an average video length of 4.6 seconds. OVIS contains 901 videos and corresponding annotations of \textit{occluded objects} (\ie masks of occluded objects are also labelled). OVIS contains 901 videos, 25 categories, 296k masks, an average of 5.8 instances per video and an average video length of 10.1 seconds. Both datasets are manually labelled every 5 frames.

\noindent \textbf{Dataset bias.} Figure~\ref{fig:inst_vos_dataset} (left) shows the static and dynamic biases of a VisTR-R50 model trained on both datasets using the layer-wise metric, (Eq.~\ref{eq:biasscores}). The model trained on OVIS contains notably more dynamic bias than the YTVIS trained model in the last layer of the encoder (\ie layer six) while the biases are similar in the final decoder layer (\ie layer 12). This result is further demonstrated when observing the unit-wise results, (Eq.~\ref{eq:bias_scores_indv}), shown in Fig.~\ref{fig:inst_vos_dataset} (right). While both datasets produce solely static and joint units, the model trained on OVIS produces more jointly encoding units in the encoder than the model trained on YTVIS and the opposite is true in the decoder. 
These results suggest that while dynamics are learned in the encoding layers, the model may lack the ability to decode the dynamics. 
We observe a similar pattern when comparing to the biases of action recognition transformers (Sec.~\ref{sec:ar_models}) where solely static and joint units also are produced. This pattern might be due to the mixing of information that is present in self-attention layers, which could inhibit specialized dynamic units from emerging. An interesting future direction suggested by these results is the design of self-attention layers which can admit specialized dynamic units throughout the model (\eg by introducing two streams, as in the SlowFast model). The unit-wise results show that OVIS has a similar pattern to TAO-VOS: When moving deeper through the model, less dynamic and joint units are produced while more static units emerge. Contrastingly, all other AVOS datasets and YTVIS produce notably more joint units in the final layer. This pattern suggests that OVIS may be a better dataset for pre-training if the goal is for the model to encode maximal dynamics.




\subsubsection{Summary and shared insights}
For action recognition, our results raise questions about some of the widely adopted datasets. In particular, Diving48 is claimed to be a good benchmark for learning dynamics~\cite{li2018resound}, while our results suggest that SSv2 is better suited for evaluating a model's ability to capture dynamics. In AVOS, we found training on TAO-VOS yields the largest number of specialized dynamic units. Thus, it may be a better training dataset for tasks that rely on capturing dynamics (\eg motion segmentation). Similarly, for VIS, we demonstrated that OVIS may provide a better signal for learning dynamic information, particularly in the earlier layers of a network.

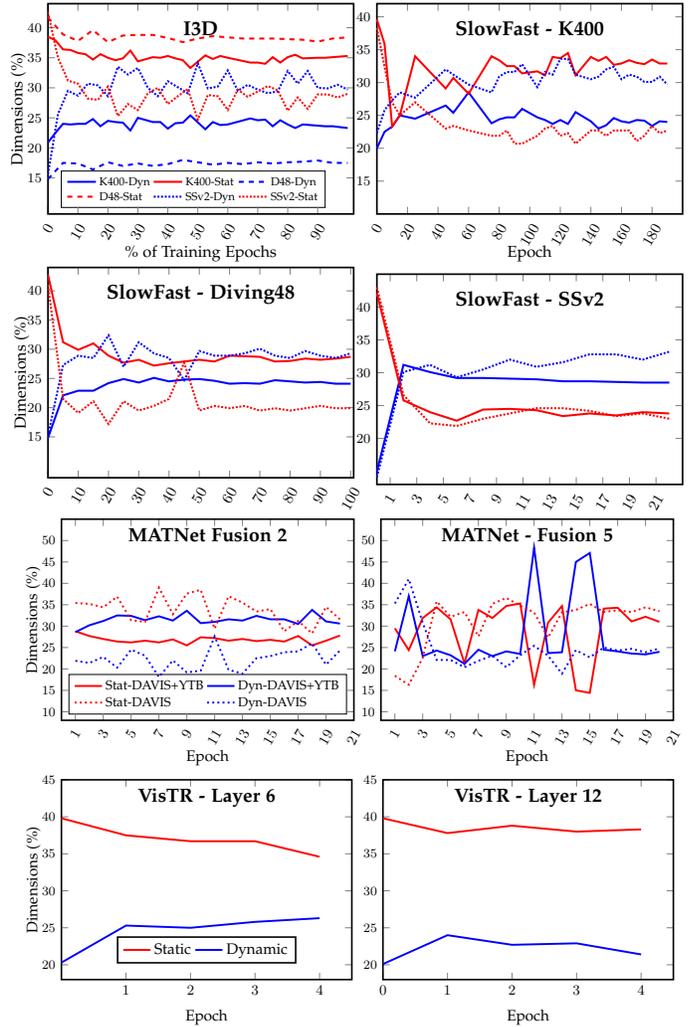
\begin{figure} [t]
\begin{center}
     \centering 
		\resizebox{0.47\textwidth}{!}{
		\begin{tikzpicture}[trim axis left,trim axis right]\ref{legend_i3d}
    \begin{axis}[
      line width=1.0,
        title={\textbf{I3D}},
        title style={at={(axis description cs:0.5,0.9)},anchor=north,font=\large},
        xlabel={\% of Training Epochs},
        x label style={font=\small},
        ylabel={Dimensions (\%)},
        xmin=0, xmax=102,
        ymin=9, ymax=44,
        xtick={0,10,20,30,40,50,60,70,80,90},
        ytick={15, 20, 25, 30, 35, 40},
        x tick label style={font=\small, rotate=60},
        y tick label style={font=\footnotesize},
        x label style={at={(axis description cs:0.5,0)},anchor=north,font=\normalsize},
        y label style={at={(axis description cs:.14,.5)},anchor=south,font=\normalsize},
        width=8cm,
        height=6cm,        
        ymajorgrids=false,
        xmajorgrids=false,
        major grid style={dotted,green!20!black},
        legend style={
         legend columns=3,
         nodes={scale=0.75, transform shape},
         cells={anchor=west},
         legend style={at={(3.05,0.13)},anchor=south,row sep=0.01pt}, font =\small},
         legend entries={[black]K400-Dyn,[black]K400-Stat,[black]D48-Dyn,[black]D48-Stat,[black]SSv2-Dyn,[black]SSv2-Stat},
        legend to name=legend_i3d,
    ]
    \addplot[line width=1.2pt,  mark options={line width=0.5pt,scale=1.2,solid}, color=blue, style=solid]
        coordinates {(0,20.9)(2.5,22.5)(5,24.0)(7.5,23.9)(10,24.0)(12.5,24.0)(15,24.8)(17.5,23.6)(20,24.5)(22.5,24.3)(25,24.2)(27.5,22.9)(30,25.0)(35,24.3)(37.5,24.3)(40,23.2)(42.5,24.1)(45,24.2)(47.5,25.4)(52.5,23.1)(55,24.3)(57.5,23.8)(60,23.9)(67.5,24.9)(70,24.6)(72.5,24.7)(75,23.6)(77.5,24.6)(80,23.9)(82.5,23.3)(85,23.9)(90,23.7)(92.5,23.6)(95,23.6)(100,23.3)};
    \addplot[line width=1.2pt,  mark options={line width=0.5pt,scale=1.2,solid}, color=red, style=solid]
        coordinates {(0,38.5)(2.5,37.9)(5,36.4)(7.5,36.3)(10,35.8)(12.5,35.6)(15,34.7)(17.5,35.6)(20,35.0)(22.5,34.6)(25,34.9)(27.5,36.2)(30,34.4)(35,35.1)(37.5,35.0)(40,35.3)(42.5,34.9)(45,34.6)(47.5,33.3)(52.5,35.4)(55,34.8)(57.5,35.3)(60,35.0)(67.5,34.2)(70,34.2)(72.5,34.0)(75,35.1)(77.5,34.2)(80,35.1)(82.5,35.5)(85,34.9)(90,35.0)(92.5,35.0)(95,35.1)(100,35.3)};
        
    \addplot[line width=1.2pt,  mark options={line width=0.5pt,scale=1.2,solid}, color=blue, style=dashed]
        coordinates {(0,14.8)(5,17.5)(10,17.4)(15,16.4)(20,17.6)(25,17.0)(30,17.4)(35,17.0)(40,17.3)(45,18.0)(50,17.6)(55,17.2)(60,17.5)(65,17.3)(70,17.6)(75,17.4)(80,17.6)(85,17.7)(90,17.9)(95,17.5)(100,17.5)};
    \addplot[line width=1.2pt,  mark options={line width=0.5pt,scale=1.2,solid}, color=red, style=dashed]
        coordinates {(0,41.7)(5,38.9)(10,37.8)(15,39.6)(20,37.6)(25,38.7)(30,38.8)(35,38.8)(40,38.3)(45,37.6)(50,38.2)(55,38.6)(60,38.4)(65,38.2)(70,38.2)(75,38.2)(80,38.1)(85,38.0)(90,37.7)(95,38.2)(100,38.4)};

    \addplot[line width=1.2pt,  mark options={line width=0.5pt,scale=1.2,solid}, color=blue, style=densely dotted]
        coordinates {(0,15.6)(3.33,25.6)(6.66,29.5)(10,28.7)(13.33,30.7)(16.66,30.4)(20,28.6)(23.33,33.5)(26.66,32.2)(30,33.3)(33.33,30.0)(36.66,28.6)(40,31.1)(43.33,30.1)(46.66,29.1)(50,34.1)(53.33,29.9)(56.66,30.2)(60,32.9)(63.33,29.5)(66.66,30.5)(70,29.7)(73.33,29.1)(76.66,29.4)(80,32.8)(83.33,30.7)(86.66,32.8)(90,30.0)(93.33,29.9)(96.66,30.5)(100,29.7)};
    \addplot[line width=1.2pt,  mark options={line width=0.5pt,scale=1.2,solid}, color=red, style=densely dotted]
        coordinates {(0,42.6)(3.33,35.0)(6.66,31.2)(10,30.7)(13.33,28.2)(16.66,28.0)(20,30.2)(23.33,25.3)(26.66,27.3)(30,26.3)(33.33,29.1)(36.66,30.0)(40,27.3)(43.33,28.6)(46.66,29.8)(50,24.7)(53.33,28.8)(56.66,28.6)(60,26.3)(63.33,29.7)(66.66,28.5)(70,29.4)(73.33,30.3)(76.66,29.6)(80,26.2)(83.33,28.4)(86.66,26.5)(90,28.9)(93.33,28.9)(96.66,28.4)(100,29.0)};
    
\end{axis}

\end{tikzpicture}
\hspace{0.3cm}
\begin{tikzpicture}[trim axis left,trim axis right]\ref{two_stream_leg}
    \begin{axis}[
      line width=1.0,
        title={\textbf{SlowFast - K400}},
        title style={at={(axis description cs:0.5,0.9)},anchor=north,font=\large},
        xlabel={Epoch},
        x label style={font=\small},
        xmin=0, xmax=200,
        ymin=10, ymax=42,
        xtick={0,20,40,60,80,100,120,140,160,180},
        ytick={15, 20, 25, 30, 35, 40},
        x tick label style={font=\small, rotate=60},
        y tick label style={font=\footnotesize},
        x label style={at={(axis description cs:0.5,0)},anchor=north,font=\normalsize},
        width=8cm,
        height=6cm,        
        ymajorgrids=false,
        xmajorgrids=false,
    ]

    \addplot[line width=1.2pt,mark size=1.1pt,color=blue]
        coordinates {(0,20.0)(5,22.5)(10,23.3)(15,25.0)(25,24.5)(45,26.5)(50,25.4)(60,28.5)(75,23.8)(80,24.4)(85,24.7)(90,24.7)(95,26.0)(105,24.7)(110,24.3)(115,23.7)(120,24.3)(125,23.7)(130,25.5)(140,24.1)(145,23.0)(150,23.5)(155,24.6)(160,24.1)(165,23.9)(170,24.3)(175,24.2)(180,23.4)(185,24.1)(190,24.0)};
    \addplot[line width=1.2pt,mark size=1.1pt,color=red]
        coordinates {(0,39.6)(5,35.9)(10,23.3)(15,25.0)(25,34.0)(45,29.1)(50,30.7)(60,28.3)(75,34.0)(80,33.4)(85,32.5)(90,32.5)(95,31.4)(105,31.7)(110,31.1)(115,33.9)(120,33.7)(125,34.5)(130,31.1)(140,33.9)(145,33.3)(150,33.9)(155,32.7)(160,32.9)(165,33.4)(170,33.0)(175,32.8)(180,33.5)(185,32.9)(190,32.9)};
        
    \addplot[densely dotted,line width=1.2pt,mark size=1.1pt,color=blue]
        coordinates {(0,22.3)(5,25.8)(10,27.3)(15,28.5)(25,27.7)(45,32.0)(50,31.2)(60,29.7)(75,28.5)(80,30.9)(85,31.6)(90,31.6)(95,32.8)(105,29.3)(110,31.6)(115,31.2)(120,33.6)(125,33.6)(130,31.2)(140,30.5)(145,30.9)(150,32.0)(155,32.4)(160,30.5)(165,31.2)(170,30.9)(175,30.1)(180,30.1)(185,30.9)(190,29.7)};
     \addplot[densely dotted,line width=1.2pt,mark size=1.1pt,color=red]
        coordinates {(0,38.3)(5,32.4)(10,27.0)(15,25.0)(25,27.0)(45,23.0)(50,23.4)(60,22.7)(75,21.9)(80,21.9)(85,22.7)(90,20.7)(95,20.7)(105,21.9)(110,23.0)(115,23.4)(120,21.9)(125,22.3)(130,20.7)(140,22.7)(145,22.7)(150,21.9)(155,22.7)(160,22.7)(165,22.7)(170,21.1)(175,21.9)(180,23.4)(185,22.3)(190,22.7)};
\end{axis}

\end{tikzpicture}
}
\resizebox{0.47\textwidth}{!}{
\begin{tikzpicture}[trim axis left,trim axis right]\ref{two_stream_leg}
    \begin{axis}[
      line width=1.0,
        title={\textbf{SlowFast - Diving48}},
        title style={at={(axis description cs:0.5,0.9)},anchor=north,font=\large},
        x label style={font=\small},
        xmin=0, xmax=101,
        ymin=8, ymax=44,
        xtick={0,10,20,30,40,50,60,70,80,90,100},
        ytick={15,20,25,30,35,40},
        ylabel={Dimensions (\%)},
        x tick label style={font=\small, rotate=60},
        y tick label style={font=\footnotesize},
        x label style={at={(axis description cs:0.5,0)},anchor=north,font=\normalsize},
        y label style={at={(axis description cs:.15,.5)},anchor=south,font=\normalsize},
        width=8cm,
        height=6cm,        
        ymajorgrids=false,
        xmajorgrids=false,
    ]

    \addplot[line width=1.2pt,mark size=1.1pt,color=blue]
        coordinates {(0,14.8)(5,22.1)(10,22.9)(15,22.9)(20,24.2)(25,24.9)(30,24.3)(35,25.1)(40,24.5)(45,24.8)(50,24.9)(55,24.6)(60,24.1)(65,24.2)(70,24.1)(75,24.7)(80,24.5)(85,24.3)(90,24.4)(95,24.1)(100,24.1)};
    \addplot[line width=1.2pt,mark size=1.1pt,color=red]
        coordinates {(0,42.8)(5,31.2)(10,29.9)(15,31.0)(20,28.9)(25,27.7)(30,28.2)(35,27.2)(40,27.6)(45,27.9)(50,28.2)(55,27.9)(60,28.9)(65,28.8)(70,28.7)(75,27.9)(80,28.0)(85,28.4)(90,28.2)(95,28.4)(100,28.7)};

    \addplot[densely dotted,line width=1.2pt,mark size=1.1pt,color=blue]
        coordinates {(0,15.2)(5,27.3)(10,28.9)(15,28.5)(20,32.4)(25,27.0)(30,31.2)(35,29.3)(40,28.5)(45,24.7)(50,29.7)(55,28.9)(60,28.9)(65,29.3)(70,30.1)(75,28.9)(80,28.5)(85,29.7)(90,28.9)(95,28.5)(100,29.3)};
     \addplot[densely dotted,line width=1.2pt,mark size=1.1pt,color=red]
        coordinates {(0,41.0)(5,21.5)(10,19.1)(15,21.1)(20,17.2)(25,21.1)(30,19.5)(35,20.3)(40,21.5)(45,27.7)(50,19.5)(55,20.3)(60,19.9)(65,20.3)(70,19.5)(75,19.9)(80,19.5)(85,19.9)(90,20.3)(95,19.9)(100,19.9)};
\end{axis}
\end{tikzpicture}
\hspace{0.3cm}
\begin{tikzpicture}[trim axis left,trim axis right]\ref{two_stream_leg}
    \begin{axis}[
      line width=1.0,
        title={\textbf{SlowFast - SSv2}},
        title style={at={(axis description cs:0.5,0.9)},anchor=north,font=\large},
        x label style={font=\small},
        xmin=0, xmax=23,
        ymin=13, ymax=45,
        xtick={1,3,5,7,9,11,13,15,17,19,21},
        ytick={20, 25, 30, 35, 40},
        x tick label style={font=\small, rotate=60},
        y tick label style={font=\footnotesize},
        x label style={at={(axis description cs:0.5,0)},anchor=north,font=\normalsize},
        width=8cm,
        height=6cm,        
        ymajorgrids=false,
        xmajorgrids=false,
        legend style={
		 legend columns=2,
         nodes={scale=0.85, transform shape},
         cells={anchor=west},
         legend style={at={(3,0.13)},anchor=south,row sep=0.01pt}, font =\normalsize},
         legend entries={[black]DynSlow,[black]StatSlow,[black]DynFast,[black]StatFast},
        legend to name=two_stream_leg,
    ]

    \addplot[line width=1.2pt,mark size=1.1pt,color=blue]
        coordinates {(0,15.0)(2,31.2)(4,30.1)(6,29.2)(8,29.2)(10,29.1)(12,29.0)(14,28.7)(16,28.7)(18,28.6)(20,28.5)(22,28.5)};
    \addplot[line width=1.2pt,mark size=1.1pt,color=red]
        coordinates {(0,42.1)(2,25.8)(4,24.0)(6,22.7)(8,24.4)(10,24.5)(12,24.3)(14,23.4)(16,23.8)(18,23.5)(20,24.0)(22,23.8)};
        
    \addplot[densely dotted,line width=1.2pt,mark size=1.1pt,color=blue]
        coordinates {(0,14.1)(2,30.1)(4,31.2)(6,29.3)(8,30.5)(10,32.0)(12,30.9)(14,31.6)(16,32.8)(18,32.8)(20,32.0)(22,33.2)};
     \addplot[densely dotted,line width=1.2pt,mark size=1.1pt,color=red]
        coordinates {(0,43.0)(2,26.6)(4,22.3)(6,21.9)(8,23.0)(10,23.8)(12,24.6)(14,24.6)(16,24.2)(18,23.4)(20,23.8)(22,23.0)};
\end{axis}

\end{tikzpicture}
}
\resizebox{0.45\textwidth}{!}{
\begin{tikzpicture}[trim axis left,trim axis right] \ref{vos_dataset_per_epoch_legend}
\begin{axis}[
      line width=1.0,
        title={\textbf{MATNet Fusion 2}},
        title style={at={(axis description cs:0.5,0.94)},anchor=north,font=\large},
        xlabel={Epoch},
        x label style={font=\small},
        ylabel={Dimensions (\%)},
        xmin=0, xmax=21,
        ymin=8, ymax=55,
        xtick={1,3,5,7,9,11,13,15,17,19,21},
        ytick={10, 15, 20, 25, 30, 35, 40, 45, 50},
         x tick label style={font=\small, rotate=60},
        y tick label style={font=\footnotesize},
        x label style={at={(axis description cs:0.5,0)},anchor=north,font=\normalsize},
        y label style={at={(axis description cs:.14,.5)},anchor=south,font=\normalsize},
        width=8cm,
        height=6cm,        
        ymajorgrids=false,
        xmajorgrids=false,
        legend style={
		 legend columns=2,
         nodes={scale=0.85, transform shape},
         cells={anchor=west},
         legend style={at={(3.21,0.13)},anchor=south,row sep=0.01pt}, font =\normalsize},
         legend entries={[black]Stat-DAVIS+YTB,[black]Dyn-DAVIS+YTB,[black]Stat-DAVIS,[black]Dyn-DAVIS},
        legend to name=vos_dataset_per_epoch_legend,
    ]
    \addplot[line width=1.2pt,mark size=1.1pt,color=red]
        coordinates{(1,28.8)(2,27.7)(3,27.0)(4,26.4)(5,26.2)(6,26.6)(7,26.2)(8,26.9)(9,25.5)(10,27.4)(11,27.2)(12,26.6)(13,27.0)(14,26.5)(15,26.8)(16,26.4)(17,27.7)(18,25.5)(19,26.6)(20,27.8)};
   \addplot[line width=1.2pt,mark size=1.1pt,color=blue]
            coordinates{(1,28.6)(2,30.2)(3,31.2)(4,32.5)(5,32.4)(6,31.4)(7,32.3)(8,31.3)(9,33.6)(10,30.7)(11,31.0)(12,31.6)(13,31.3)(14,32.4)(15,31.6)(16,31.6)(17,30.5)(18,33.8)(19,31.1)(20,30.6)};
    
    \addplot[line width=1.2pt,mark size=1.1pt,color=red,style=dotted]
    coordinates{(1,35.4)(2,35.2)(3,34.4)(4,36.9)(5,31.4)(6,31.0)(7,39.0)(8,32.8)(9,37.6)(10,38.5)(11,29.4)(12,37.0)(13,35.5)(14,33.4)(15,34.0)(16,28.8)(17,31.2)(18,28.3)(19,34.5)(20,31.6)};
   \addplot[line width=1.2pt,mark size=1.1pt,color=blue,style=dotted]
    coordinates {(1,21.9)(2,21.4)(3,22.8)(4,20.3)(5,24.5)(6,23.1)(7,18.1)(8,22.0)(9,19.2)(10,19.6)(11,27.6)(12,19.8)(13,18.9)(14,22.5)(15,23.0)(16,23.9)(17,24.1)(18,26.0)(19,21.0)(20,24.3)};
     
\end{axis}
\end{tikzpicture}%
\hspace{0.5cm}
\begin{tikzpicture}[trim axis left,trim axis right]
  \begin{axis}[
      line width=1.0,
        title={\textbf{MATNet - Fusion 5}},
        title style={at={(axis description cs:0.5,0.94)},anchor=north,font=\large},
        xlabel={Epoch},
        x label style={font=\small},
        xmin=0, xmax=21,
        ymin=8, ymax=55,
        xtick={1,3,5,7,9,11,13,15,17,19,21},
        ytick={10, 15, 20, 25, 30, 35, 40, 45, 50},
         x tick label style={font=\small, rotate=60},
        y tick label style={font=\footnotesize},
        x label style={at={(axis description cs:0.5,0)},anchor=north,font=\normalsize},
        width=8cm,
        height=6cm,        
        ymajorgrids=false,
        xmajorgrids=false,
    ]
     \addplot[line width=1.2pt,mark size=1.1pt,color=red]
            coordinates{(1,29.5)(2,24.4)(3,31.9)(4,34.4)(5,31.6)(6,21.3)(7,33.8)(8,31.9)(9,34.7)(10,35.3)(11,16.3)(12,30.8)(13,34.7)(14,15.0)(15,14.4)(16,34.1)(17,34.3)(18,31.1)(19,32.2)(20,31.0)};
            
    \addplot[line width=1.2pt,mark size=1.1pt,color=blue]
            coordinates{(1,24.1)(2,37.0)(3,23.0)(4,24.3)(5,23.2)(6,21.2)(7,24.5)(8,23.0)(9,24.1)(10,23.5)(11,48.3)(12,23.7)(13,23.9)(14,45.0)(15,47.1)(16,24.5)(17,24.1)(18,23.6)(19,23.4)(20,24.0)};
            
    \addplot[line width=1.2pt,mark size=1.1pt,color=red,style=dotted]
        coordinates {(1,18.4)(2,16.3)(3,22.6)(4,35.8)(5,32.2)(6,33.3)(7,27.6)(8,35.2)(9,36.6)(10,34.8)(11,33.2)(12,27.3)(13,33.2)(14,33.8)(15,35.2)(16,33.3)(17,33.7)(18,33.3)(19,34.4)(20,33.5)};
    \addplot[line width=1.2pt,mark size=1.1pt,color=blue,style=dotted]
        coordinates {(1,35.3)(2,41.0)(3,30.8)(4,22.1)(5,22.1)(6,20.4)(7,21.9)(8,23.1)(9,20.4)(10,23.2)(11,25.4)(12,23.1)(13,18.9)(14,24.4)(15,22.7)(16,25.0)(17,24.3)(18,24.7)(19,23.9)(20,24.8)};
    
\end{axis}
\end{tikzpicture}}
\resizebox{0.45\textwidth}{!}{
    \begin{tikzpicture}[trim axis left,trim axis right]
  \begin{axis}[
      line width=1.0,
        title={\textbf{VisTR - Layer 6}},
        title style={at={(axis description cs:0.5,0.94)},anchor=north,font=\large},
        xlabel={Epoch},
        x label style={font=\small},
        ylabel={Dimensions (\%)},
        xmin=0, xmax=4.5,
        ymin=18, ymax=45,
        xtick={1,2,3,4},
        ytick={10, 15, 20, 25, 30, 35, 40, 45, 50},
         x tick label style={font=\small, rotate=0},
        y tick label style={font=\footnotesize},
        x label style={at={(axis description cs:0.5,0)},anchor=north,font=\normalsize},
        y label style={at={(axis description cs:.14,.5)},anchor=south,font=\normalsize},
        width=8cm,
        height=6cm,        
        ymajorgrids=false,
        xmajorgrids=false,
        major grid style={dotted,green!20!black},
        legend style={
		 legend columns=2,
          legend style={at={(0.5,0.08)},anchor=south,row sep=0.01pt}}
    ]
    \addlegendentry{Static}
     \addplot[line width=1.2pt,mark size=1.1pt,color=red]
            coordinates{(0,39.8)(1,37.5)(2,36.7)(3,36.7)(4,34.6)};
            
    \addlegendentry{Dynamic}
    \addplot[line width=1.2pt,mark size=1.1pt,color=blue]
            coordinates{(0,20.3)(1,25.3)(2,25.0)(3,25.8)(4,26.3)};
    
\end{axis}
\end{tikzpicture}
    \hspace{0.5cm}
\begin{tikzpicture}[trim axis left,trim axis right]
  \begin{axis}[
      line width=1.0,
        title={\textbf{VisTR - Layer 12}},
        title style={at={(axis description cs:0.5,0.94)},anchor=north,font=\large},
        xlabel={Epoch},
        x label style={font=\small},
        xmin=0, xmax=4.5,
        ymin=18, ymax=45,
        xtick={0,1,2,3,4},
        ytick={10, 15, 20, 25, 30, 35, 40, 45, 50},
         x tick label style={font=\small, rotate=0},
        y tick label style={font=\footnotesize},
        x label style={at={(axis description cs:0.5,0)},anchor=north,font=\normalsize},
        width=8cm,
        height=6cm,        
        ymajorgrids=false,
        xmajorgrids=false,
        legend style={
		 legend columns=2,
         nodes={scale=0.85, transform shape},
         cells={anchor=west},
         legend style={at={(3,0.13)},anchor=south,row sep=0.01pt}, font =\normalsize},
        legend to name=two_stream_leg,
    ]
     \addplot[line width=1.2pt,mark size=1.1pt,color=red]
            coordinates{(0,39.8)(1,37.8)(2,38.8)(3,38.0)(4,38.3)};
            
    \addplot[line width=1.2pt,mark size=1.1pt,color=blue]
            coordinates{(0,20.1)(1,24.0)(2,22.7)(3,22.9)(4,21.4)};
    
\end{axis}
\end{tikzpicture}}
\end{center}
\vspace{-0.55cm}
	\caption{\textbf{Top and second row:} The encoding of static and dynamic bias for I3D~\cite{carreira2017quo} (8x8 R18) and SlowFast~\cite{feichtenhofer2019slowfast} (4x16 R18) over the course of training on Kinetics-400. \textbf{Third row} The encoding of static and dynamic bias over the course of training for MATNet trained on DAVIS and DAVIS + YouTube-VOS. \textbf{Fourth row} The encoding of static and dynamic bias for the VisTR~\cite{wang2021end} model over the course of training on the OVIS dataset~\cite{qi2022occluded}.}\label{fig:per_epoch}
 \vspace{-0.6cm}
\end{figure}

\vspace{-0.3cm}
\subsection{When are statics and dynamics learned in training?}
\vspace{-0.08cm}
\label{sec:epoch_effect}
Based on our previous analysis of models post-training, we now ask: ``What are characteristics of learning static and dynamic information over the course of training?''. Understanding how the models learn static and dynamic information can potentially inform researchers to build better training protocols for learning specific types of information.

\vspace{-0.3cm}
\subsubsection{Action Recognition}
We examine both a single and multi-stream architecture and choose the I3D~\cite{carreira2017quo} and SlowFast~\cite{feichtenhofer2019slowfast} models. We evaluate these models on Kinetics~\cite{carreira2017quo}, Diving48~\cite{li2018resound} and SSv2~\cite{goyal2017something} as each dataset differs in the static and dynamic information learned during training (see Sec.~\ref{sec:ar_dataset}). Figure~\ref{fig:per_epoch} (top and second row) shows the layer-wise results for action recognition. Both the static and dynamic information are largely learned during the first half of training in all cases. Comparing both models for Kinetics-400 and Diving48 further confirms our previous finding (see Sec.~\ref{sec:ar_dataset}) that both datasets result in similar amounts of static and dynamic information. Conversely, models trained on SSv2 encode more dynamic information for both the single stream and two-stream models. The I3D model has a balance of static and dynamic information throughout training, while both branches in the SlowFast model quickly converge to be biased toward encoding dynamic information. The latter indicates that, given a model which has the ability to encode either type of information (\eg within the two branches in the SlowFast model), SSv2 will yield a greater amount of dynamics.

\vspace{-0.2cm}
\subsubsection{Automatic video object segmentation}
\label{sec:epochwise_vos}
We train MATNet two datasets: (i) DAVIS16 only (DAVIS) and (ii) DAVIS16 combined with YouTube-VOS (DAVIS+YTB) as originally proposed~\cite{zhou2020motion}. Figure~\ref{fig:per_epoch} (third row) shows the epoch-wise biases learned by MATNet on both settings. When trained on DAVIS+YTB, fusion layer 2 learns more dynamics compared to DAVIS alone, which confirms our previous findings (see Fig.~\ref{fig:vos_dataset_indiv} fusion layer 2). Additionally, when trained on DAVIS+YTB, the model quickly converges toward being dynamics biased. On the other hand, fusion layer 5 shows an interesting pattern of alternation between the static and dynamic factors throughout training and culminates to being static biased. These results align with our previous findings that MATNet's early fusion layers converge to being dynamics biased, unlike late fusion layers (see Fig.~\ref{fig:stagewise_vos}c). This result may stem from the fact that deeper layers tend to capture more abstract information compared to earlier layers. When trained on DAVIS, the final fusion layer becomes static biased in the early epochs, which aligns with the findings from action recognition. 

\subsubsection{Video instance segmentation}
\label{sec:epochwise_vis}
Figure~\ref{fig:per_epoch} (fourth row) presents results for the VisTR model~\cite{wang2021end} trained on the OVIS dataset~\cite{qi2022occluded} for the last layer of the encoder (layer six) and decoder (layer 12). Interestingly, we observe different patterns in each layer. In layer six, the dynamic bias increases monotonically during training while static bias decreases over training. In layer 12, the dynamic bias increases by $>4\%$ in the first epoch but decreases thereafter. This pattern suggests that the encoder and decoder have different training dynamics as they converge to their resulting biases in alternate ways. These results align with the results from Figure~\ref{fig:inst_vos_arch} that also show differences between the encoder and decoder in terms of the ratio of joint to static units.

\vspace{-0.2cm}
\subsubsection{Summary and shared insights}
All models in the three tasks converge to their culminating biases within the first half of training epochs, except for MATNet trained on DAVIS+YTB. We also observed an interaction between models and datasets in which certain combinations of the two produce significantly more stability in their learned biases than others over the course of training.

\section{Controlling Model Bias}\label{sec:app}


We have demonstrated the effect to which architectures, datasets and training protocols have on a model's biases. 
We now ask: (i) \textit{Is it possible to control the static and dynamic biases of a model?} and (ii) \textit{What impact do static and dynamic units have on performance?} 
We first show that the type of bias encoded by a unit greatly determines the impact it has on final performance through \textit{neuron removal} experiments for action recognition and video segmentation. 
Motivated by the varying impact different biases have on overall performance, we aim to improve the performance for each task. 
For action recognition, we propose \textit{StaticDropout}, a novel dropout strategy that uses our estimation technique to dropout static-biased units during training, with the ultimate goal of debiasing models away from static information. 
For AVOS, we perform a detailed analysis on cross connections and fusion design choices with respect to their static and dynamic biases and show how to encourage a previously static biased model to become more biased toward dynamics. In both domains, we show an improvement in performance for tasks that require dynamics.

\subsection{Neuron Removal}\label{sec:neuron_removal}
\vspace{-0.1cm}
To evaluate the effect of static and dynamic units on overall performance, we conduct perturbation experiments where we remove the top-$k$ units (\ie channels) that are biased toward the static or dynamic factor during inference and evaluate the performance drop. The removal is done by setting all activations to zero in the identified channels. We compare these static or dynamic biased units with respect to randomly selected channels. A lower Area Under Curve (AUC) suggests that the units are more important for model performance.

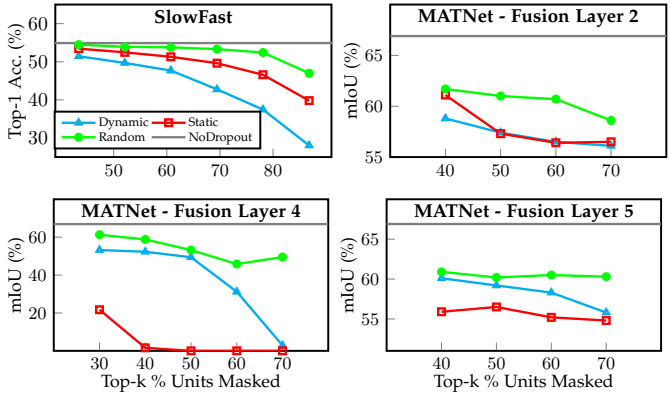
\begin{figure} [t]
	\begin{center}
     \centering 
     \resizebox{0.49\textwidth}{!}{
\begin{tikzpicture} 
                 \begin{axis}[
                 line width=1.0,
                 title={\textbf{SlowFast}},
                 title style={at={(axis description cs:0.5,0.94)},anchor=north,font=\normalsize},
                 ylabel={Top-1 Acc. (\%)},
                 xmin=39, xmax=91,
                 ymin=25, ymax=65,
                 xtick pos=left,
                 ytick pos=bottom,
                 xtick={50,60,70,80},
                 x tick label style={font=\small, rotate=0, anchor=north},
                 x label style={at={(axis description cs:0.5,0.09)},anchor=north,font=\small},
                 y label style={at={(axis description cs:0.17,.5)},anchor=south,font=\small},
                 width=6.5cm,
                 height=4.3cm, 
                 ymajorgrids=false,
                 xmajorgrids=false,
                 major grid style={dotted,green!20!black},
                 legend columns=2,
                 legend style={
                  nodes={scale=0.52, transform shape},
                  cells={anchor=west},
                  legend style={at={(0.37,0.05)},anchor=south,row sep=0.01pt}, font =\large}
             ]
             \addlegendentry{Dynamic}
             \addplot[line width=1.2pt, mark size=1.7pt, color=cyan, mark=triangle,error bars/.cd, y dir=both, y explicit,]
                     coordinates {(43.4, 51.44)
                                  (52.08, 49.69)
                                  (60.76, 47.69)
                                  (69.44, 42.77)
                                  (78.1, 37.46)
                                  (86.80, 28.05)};
            \addlegendentry{Static}
            \addplot[line width=1.2pt, mark size=1.7pt, color=red, mark=square,error bars/.cd, y dir=both, y explicit,]
                     coordinates {(43.4, 53.436)
                                  (52.08, 52.469)
                                  (60.76, 51.293)
                                  (69.44, 49.62)
                                  (78.1, 46.587)
                                  (86.80, 39.799)};
            \addlegendentry{Random}
            \addplot[line width=1.2pt, mark size=1.7pt, color=green, mark=*,error bars/.cd, y dir=both, y explicit,]
                     coordinates {(43.4, 54.48)
                                  (52.08, 53.916)
                                  (60.76, 53.775)
                                  (69.44, 53.316)
                                  (78.1, 52.415)
                                  (86.80, 46.983)
                                  };
            \addlegendentry{NoDropout}
             \addplot[line width=1.2pt, mark size=0, color=gray, mark=diamond,error bars/.cd, y dir=both, y explicit,]
                     coordinates {(0, 54.9)
                                  (100, 54.9)};
              \end{axis}
\end{tikzpicture}
\begin{tikzpicture} 
                 \begin{axis}[
                 line width=1.0,
                 title={\textbf{MATNet - Fusion Layer 2}},
                 title style={at={(axis description cs:0.5,0.94)},anchor=north,font=\normalsize},
                 ylabel={mIoU (\%)},
                 xmin=30, xmax=80,
                 ymin=55, ymax=70,
                 xtick pos=left,
                 ytick pos=bottom,
                 xtick={40,50,60,70},
                 ytick={55, 60, 65},
                 x tick label style={font=\small, rotate=0, anchor=north},
                 y tick label style={font=\small},
                 y label style={at={(axis description cs:0.17,.5)},anchor=south,font=\small},
                 width=6.5cm,
                 height=4.3cm, 
                 ymajorgrids=false,
                 xmajorgrids=false,
                 major grid style={dotted,green!20!black},
             ]

             \addplot[line width=1.2pt, mark size=1.7pt, color=cyan, mark=triangle,error bars/.cd, y dir=both, y explicit,]
                     coordinates {(40, 58.8)
                                  (50, 57.4)
                                  (60, 56.5)
                                  (70, 56.1)};
            \addplot[line width=1.2pt, mark size=1.7pt, color=red, mark=square,error bars/.cd, y dir=both, y explicit,]
                     coordinates {(40, 61.1)
                                  (50, 57.3)
                                  (60, 56.4)
                                  (70, 56.5)};
            \addplot[line width=1.2pt, mark size=1.7pt, color=green, mark=*,error bars/.cd, y dir=both, y explicit,]
                     coordinates {(40, 61.7)
                                  (50, 61.0)
                                  (60, 60.7)
                                  (70, 58.6)
            };
             \addplot[line width=1.2pt, mark size=0, color=gray, mark=*,error bars/.cd, y dir=both, y explicit,]
                     coordinates {(30, 66.9)
                                  (80, 66.9)};
              \end{axis}
\end{tikzpicture}}
\vspace{-0.2cm}
\resizebox{0.49\textwidth}{!}{
\begin{tikzpicture} 
                 \begin{axis}[
                 line width=1.0,
                 title={\textbf{MATNet - Fusion Layer 4}},
                    title style={at={(axis description cs:0.5,0.94)},anchor=north,font=\normalsize},
                 xlabel={Top-k \% Units Masked},
                 ylabel={mIoU (\%)},
                 xtick pos=left,
                 ytick pos=bottom,
                 xmin=20, xmax=80,
                 ymin=0, ymax=80,
                 xtick={30,40,50,60,70},
                 ytick={20,40,60},
                 x tick label style={font=\small, rotate=0, anchor=north},
                 y tick label style={font=\small},
                 x label style={at={(axis description cs:0.5,0.09)},anchor=north,font=\small},
                 y label style={at={(axis description cs:0.17,.5)},anchor=south,font=\small},
                 width=6.5cm,
                 height=4.3cm, 
                 ymajorgrids=false,
                 xmajorgrids=false,
                 major grid style={dotted,green!20!black},
             ]
            %
             \addplot[line width=1.2pt, mark size=1.7pt, color=cyan, mark=triangle,error bars/.cd, y dir=both, y explicit,]
                     coordinates {(30, 53.2)
                                  (40, 52.3)
                                  (50, 49.4)
                                  (60, 31.2)
                                  (70, 2.9)};
            \addplot[line width=1.2pt, mark size=1.7pt, color=red, mark=square,error bars/.cd, y dir=both, y explicit,]
                     coordinates {(30, 21.7)
                                  (40, 1.6)
                                  (50, 1.969784267764861e-06)
                                  (60, 3.4545299701601246e-05)
                                  (70, 1.0245879243979795e-07)};
            \addplot[line width=1.2pt, mark size=1.7pt, color=green, mark=*,error bars/.cd, y dir=both, y explicit,]
                     coordinates {(30, 61.3)
                                  (40, 58.8)
                                  (50, 53.2)
                                  (60, 45.8)
                                  (70, 49.5)
            };
             \addplot[line width=1.2pt, mark size=0, color=gray, mark=*,error bars/.cd, y dir=both, y explicit,]
                     coordinates {(20, 66.9)
                                  (80, 66.9)};
              \end{axis}
\end{tikzpicture}
\begin{tikzpicture} 
                 \begin{axis}[
                 line width=1.0,
                 title={\textbf{MATNet - Fusion Layer 5}},
                 title style={at={(axis description cs:0.5,0.94)},anchor=north,font=\normalsize},
                 xlabel={Top-k \% Units Masked},
                 ylabel={mIoU (\%)},
                 xmin=30, xmax=80,
                 ymin=51, ymax=70,
                xtick pos=left,
                ytick pos=bottom,
                 xtick={40,50,60,70},
                 ytick={55, 60, 65},
                 x tick label style={font=\small, rotate=0, anchor=north},
                 y tick label style={font=\small},
                 x label style={at={(axis description cs:0.5,0.09)},anchor=north,font=\small},
                 y label style={at={(axis description cs:0.17,.5)},anchor=south,font=\small},
                 width=6.5cm,
                 height=4.3cm, 
                 ymajorgrids=false,
                 xmajorgrids=false,
                 major grid style={dotted,green!20!black},
             ]
            
             \addplot[line width=1.2pt, mark size=1.7pt, color=cyan, mark=triangle,error bars/.cd, y dir=both, y explicit,]
                     coordinates {(40, 60.1)
                                  (50, 59.2)
                                  (60, 58.3)
                                  (70, 55.8)};
            \addplot[line width=1.2pt, mark size=1.7pt, color=red, mark=square,error bars/.cd, y dir=both, y explicit,]
                     coordinates {(40, 55.9)
                                  (50, 56.5)
                                  (60, 55.2)
                                  (70, 54.8)};
            \addplot[line width=1.2pt, mark size=1.7pt, color=green, mark=*,error bars/.cd, y dir=both, y explicit,]
                     coordinates {(40, 60.9)
                                  (50, 60.2)
                                  (60, 60.5)
                                  (70, 60.3)
            };
             \addplot[line width=1.2pt, mark size=0, color=gray, mark=*,error bars/.cd, y dir=both, y explicit,]
                     coordinates {(30, 66.9)
                                  (80, 66.9)};
              \end{axis}
\end{tikzpicture}
}
\end{center}
\vspace{-0.57cm}
\caption{Top-$k$ unit removal results for static and dynamic factors with respect to random units on: (action recognition) the final layer of SlowFast model trained on SSv2 dataset, and (video object segmentation) three fusion layers of the MATNet model trained on DAVIS and YouTube-VOS.}\label{fig:unit_removal}
\vspace{-0.56cm}
\end{figure}

Figure~\ref{fig:unit_removal} shows the unit attribution curves~\cite{ghorbani2019towards,chefer2021generic,fel2024holistic} for action recognition and AVOS. In action recognition we evaluate the final layer in the SlowFast model trained on SSv2 and evaluate on the SSv2 validation set to report the top-1 accuracy. As can be seen in Fig.~\ref{fig:unit_removal} (top left), the dynamic factor maximally reduces the model's performance, e.g., removing 70\% of the SlowFast model dynamic units decreases the performance by 5\% more than static units, which may be because the SlowFast model encodes a significant amount of dynamic information in the fast branch and dynamics are important for the SSv2 dataset. 


We conduct similar experiments to AVOS for the early and late fusion layers of the MATNet model trained on DAVIS and YouTube-VOS. 
Empirically, we found that at intermediate layers there is a higher chance of randomly selecting units that cumulatively have a significant impact on the performance unlike the final layers, which was also found in~\cite{cheng2022deeply}. Thus, in video object segmentation, as we remove units from the intermediate fusion layers (unlike action recognition where we ablate the final layer), we randomly sample from the units that are least biased towards the dominant factor (i.e., static or dynamic) of that layer to avoid the aforementioned scenario. More specifically, the random baseline first takes the least biased units toward the significant factor of this layer (\ie dynamics in fusion layer 2, and static in other fusion layers) and then randomly samples $x\%$ units within $(x+5)\%$ of these units.

We evaluate on the moving camouflaged animals~\cite{lamdouar2020betrayed} (MoCA) dataset, where the objects of interest are camouflaged animals and hence largely indistinguishable from their backgrounds in the absence of motion, and report the mean intersection over union (mIoU) (see Sec.~\ref{sec:vos_cc_study} for more details on evaluation). The results in Fig.~\ref{fig:unit_removal} (bottom left and right) consistently demonstrate that for every fusion layer the factor with the highest impact on performance is the factor it is most biased toward, as examined earlier (Fig.~\ref{fig:stagewise_vos}c and Sec.~\ref{sec:epochwise_vos}). More specifically, fusion layer 2 is slightly dynamic biased, while fusion layers 4 and 5 are static biased, e.g., fusion layer 4 achieves 0\% when removing only 40\% of the most static units.

In both tasks, these experiments document that masking out the top-$k$ channels based on our metric can help remove biased units in the model and consequently affect its accuracy more compared with randomly selected channels. We now aim to use these insights to show possible mechanisms for improving the performance of a model.

\begin{figure}[t]
    \includegraphics[width=0.5\textwidth]{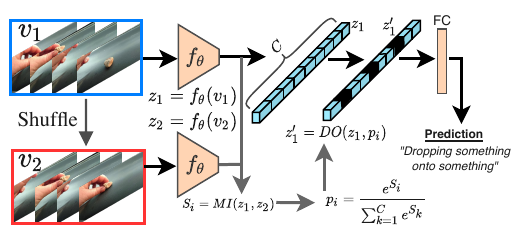}
    \vspace{-0.67cm}
    \caption{An overview of our StaticDropout method. Given a video, $v_1$, we generate $v_2$ by randomly shuffling the frames. $v_1$ and $v_2$ are passed through the model, $f_\theta$, to produce intermediate representations of the videos, $z_1$ and $z_2$. Using these representations, we estimate the mutual information (using Eq.~\ref{eq:biasscores}) of each channel, $i$, between the two representations, $S_i$. We determine the probability of dropping out channel, $p_i$, by performing a Softmax over $S_i$. }
    \label{fig:staticdropout_method}
    \vspace{-0.5cm}
\end{figure}

\vspace{-0.4cm}
\subsection{StaticDropout}\label{sec:staticdropout}
\vspace{-0.1cm}
We propose StaticDropout, a semantically guided dropout technique with the goal of static-debiasing a model. Previous work, InfoDropout~\cite{shi2020informative}, used a measure of self information to identify neurons in a CNN that encode texture information for image-based models.  Alternatively, we perform StaticDropout during training on static-biased channels in action recognition models. InfoDropout~\cite{shi2020informative} observed that dropping out neurons encoding high frequency caused the model to encode more low-frequency (e.g., shape) information. Accordingly, our intuition is that dropping out units biased towards static information will force the model to rely on \textit{dynamic} information.

\textbf{Approach.} Our overall approach is outlined in Fig.~\ref{fig:staticdropout_method}. We determine the channels in the model encoding static information by constructing statically similar video pairs during training. More specifically, given a video $v_1$, we shuffle the video frames to generate $v_2$. We pass both representations through the model to obtain intermediate representations $z_1$ and $z_2$ from layer $l$ (for fair comparison against standard dropout, we set $l$ to the last layer before the fully connected layer, and hereon omit $l$ for brevity). Using the unitwise metric, (Eq.~\ref{eq:bias_scores_indv}), we calculate a score, $S_i$, representing the static information encoded by channel $i$. We calculate the probability of dropping channel $i$ via a multinomial distribution defined as

\vspace{-0.25cm}
\begin{equation}\label{eq:static_dropout}
p_i = \frac{e^{S_i}}{\sum^C_{k=1} e^{S_k}}.
\end{equation}
The dropout rate corresponds to the total fraction of channels being dropped out of the layer, with $p_i$ the probability (relative to other channels) that channel $i$ will be included in the dropped channels.

\noindent\textbf{Evaluation Protocol.} We first demonstrate the ability of our proposed StaticDropout technique to manipulate static and dynamic biases as measured by two different metrics. For the first metric, we use the unitwise metric, (Eq.~\ref{eq:ind_bias_scores_diff_b}), to calculate the ratio of dynamic units relative to dynamic and static units. The second metric is the model's relative validation performance on shuffled \vs unshuffle frames. We apply StaticDropout to two different 3D-Resnet variants (\ie the SlowOnly architecture) with 18 and 50 layers, on two different datasets, Diving48 and SSv2.

\textbf{Implementation Details.} Models trained on SSv2 are trained with the original SlowFast~\cite{feichtenhofer2019slowfast} repository hyperparameters. 
They are trained for 30 epochs and use a cosine learning rate decay with three warmup epochs, a base learning rate of 0.1 and a warmup starting learning rate of 0.08.
Models trained on Diving48 are trained for 100 epochs and use a cosine learning rate decay with 10 warmup epochs, a base learning rate of 0.1 and a warmup starting learning rate of 0.01. We experiment with dropout rates of $r=\{0.1,0.3,0.5,0.7\}$. The static channel scores, $S_i$, are re-estimated using Eq.~\ref{eq:static_dropout} every 30 iterations. Following InfoDropout~\cite{shi2020informative}, we finetune our models without any dropout with a learning rate of 1e-5 for two and five epochs on SSv2 and Diving48, respectively. For baselines, we consider the same model architectures with standard dropout~\cite{srivastava2014dropout} and without any dropout applied.

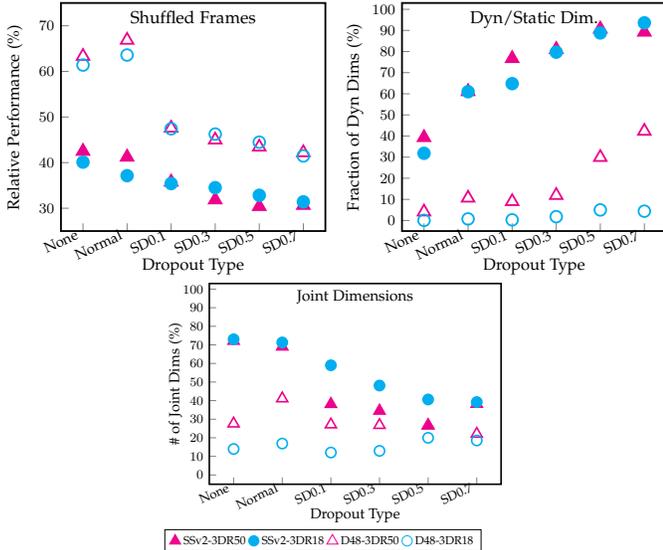
\begin{figure} [t]
	\begin{center}
     \centering 
     \resizebox{0.49\textwidth}{!}{
	\begin{tikzpicture}
    \begin{axis}[
       line width=1.0,
        title={Shuffled Frames},
        title style={at={(axis description cs:0.5,0.95)},anchor=north,font=\normalsize},
        xlabel={Dropout Type},
        ylabel={Relative Performance (\%)},
        ymin=25, ymax=75,
        ytick={0,10,20,30,40,50,60,70,80},
        symbolic x coords={None, Normal, SD0.1,SD0.3, SD0.5, SD0.7},
        xtick=data,
        x tick label style={font=\footnotesize,rotate=25,anchor=east},
        y tick label style={font=\footnotesize},
        x label style={at={(axis description cs:0.5,0.03)},anchor=north,font=\small},
        y label style={at={(axis description cs:0.12,.5)},anchor=south,font=\small},
        xtick pos=left,
        ytick pos=bottom,
        width=6.7cm,
        height=6cm,        
        ymajorgrids=false,
        xmajorgrids=false,
        major grid style={dotted,green!20!black},
    ]
    \addplot[only marks,mark size=4pt,color=magenta,mark=triangle*,]
    coordinates {(None,42.486) (Normal,41.220) (SD0.1,35.748) (SD0.3,31.851) (SD0.5,30.389) (SD0.7,30.635)};
        
    \addplot[only marks,mark size=3.2pt,color=cyan,mark=*,]
        coordinates {(None,40.113) (Normal,37.138) (SD0.1,35.407) (SD0.3,34.510) (SD0.5,32.837) (SD0.7,31.404)};
        
    \addplot[only marks,mark size=4pt,color=magenta,mark=triangle,]
    coordinates {(None,63.323) (Normal,66.849) (SD0.1,47.576) (SD0.3,44.987) (SD0.5,43.433) (SD0.7,42.193)};
        
    \addplot[only marks,mark size=3.2pt,color=cyan,mark=o,]
        coordinates {(None,61.391) (Normal,63.591) (SD0.1,47.413) (SD0.3,46.258) (SD0.5,44.479) (SD0.7,41.466)};
        
    \end{axis}
\end{tikzpicture}
\hspace{0.1cm}
	\begin{tikzpicture} 
    \begin{axis}[
       line width=1.0,
        title={Dyn/Static Dim.},
        title style={at={(axis description cs:0.5,0.95)},anchor=north,font=\normalsize},
        xlabel={Dropout Type},
        ylabel={Fraction of Dyn Dims (\%)},
        ymin=-5, ymax=103,
        ytick={0,10,20,30,40,50,60,70,80,90,100},
        symbolic x coords={None, Normal, SD0.1,SD0.3, SD0.5, SD0.7},
        xtick=data,
        x tick label style={font=\footnotesize,rotate=25,anchor=east},
        y tick label style={font=\footnotesize},
        x label style={at={(axis description cs:0.5,0.03)},anchor=north,font=\small},
        y label style={at={(axis description cs:0.12,.5)},anchor=south,font=\small},
        xtick pos=left,
        ytick pos=bottom,
        width=6.7cm,
        height=6cm,        
        ymajorgrids=false,
        xmajorgrids=false,
        major grid style={dotted,green!20!black},
        legend style={
         nodes={scale=0.9, transform shape},
         cells={anchor=west},
         legend style={at={(1.4,0.25)},anchor=south}, font =\footnotesize},
         legend entries={[black]3DResNet50,[black]3DResNet18},
        legend to name=dim_ss_legend,
    ]
    
    \addplot[only marks,mark size=4pt,color=magenta,mark=triangle*,]
    coordinates {(None,39.28571429) (Normal,61.047463180) (SD0.1,76.7) (SD0.3,81.0) (SD0.5,90.7) (SD0.7,89.2)};
        
    \addplot[only marks,mark size=3.2pt,color=cyan,mark=*,]
        coordinates {(None,31.80987203) (Normal,60.9671848) (SD0.1,64.8) (SD0.3,79.7) (SD0.5,88.8) (SD0.7,93.6)};
    
    \addplot[only marks,mark size=4pt,color=magenta,mark=triangle,]
    coordinates {(None,4.085872576) (Normal,10.64204046) (SD0.1,9.0) (SD0.3,11.9) (SD0.5,29.9) (SD0.7,42.3)};
    \addplot[only marks,mark size=3.2pt,color=cyan,mark=o,]
        coordinates {(None,0.05733944954) (Normal,0.7343941248) (SD0.1,0.3) (SD0.3,1.8) (SD0.5,5.0) (SD0.7,4.4)};
        
    \end{axis}
\end{tikzpicture}}
\resizebox{0.25\textwidth}{!}{
\begin{tikzpicture}\ref{bias_eval_legend2}
    \begin{axis}[
       line width=1.0,
        title={Joint Dimensions},
        title style={at={(axis description cs:0.5,0.95)},anchor=north,font=\normalsize},
        xlabel={Dropout Type},
        ylabel={\# of Joint Dims (\%)},
        ymin=-5, ymax=103,
        ytick={0,10,20,30,40,50,60,70,80,90,100},
        symbolic x coords={None, Normal, SD0.1,SD0.3, SD0.5, SD0.7},
        xtick=data,
        x tick label style={font=\footnotesize,rotate=25,anchor=east},
        y tick label style={font=\footnotesize},
        x label style={at={(axis description cs:0.5,0.03)},anchor=north,font=\small},
        y label style={at={(axis description cs:0.12,.5)},anchor=south,font=\small},
        xtick pos=left,
        ytick pos=bottom,
        width=8cm,
        height=6cm,        
        ymajorgrids=false,
        xmajorgrids=false,
        major grid style={dotted,green!20!black},
        legend style={
         nodes={scale=0.9, transform shape},
         cells={anchor=west},
         legend style={at={(2.5,-1.5)},anchor=south}, font =\footnotesize},
         legend columns=4,
         legend entries={[black]SSv2-3DR50,[black]SSv2-3DR18,[black]D48-3DR50,[black]D48-3DR18},
        legend to name=bias_eval_legend2,
    ]
    
    \addplot[only marks,mark size=4pt,color=magenta,mark=triangle*,]
    coordinates {(None,72.1) (Normal,69.0) (SD0.1,38.1) (SD0.3,34.5) (SD0.5,26.5) (SD0.7,38.2)};
        
    \addplot[only marks,mark size=3.2pt,color=cyan,mark=*,]
        coordinates {(None,73.0) (Normal,71.3) (SD0.1,59.0) (SD0.3,48.1) (SD0.5,40.6) (SD0.7,39.2)};
    
    \addplot[only marks,mark size=4pt,color=magenta,mark=triangle,]
    coordinates {(None,27.58789063) (Normal,41.11328125) (SD0.1,27.09960938) (SD0.3,26.80664063) (SD0.5,26.7578125) (SD0.7,22.0703125)};
    \addplot[only marks,mark size=3.2pt,color=cyan,mark=o,]
        coordinates {(None,13.96484375) (Normal,16.89453125) (SD0.1,12.06054688) (SD0.3,12.93945313) (SD0.5,19.97070313) (SD0.7,18.5546875)};
        
    \end{axis}
\end{tikzpicture}}
\end{center}
\vspace{-0.5cm}
\caption{StaticDropout influences the static and dynamic information of a model. We evaluate the relative performance of a model evaluate on normal \vs shuffled frames at validation time (left), show the fraction of channels encoding dynamic information relative to the channels encoding either static or dynamic (right), and the number of joint encoding units (bottom) for 3D-ResNet50 and 18 on the SSv2 and Diving48 datasets as a function of StaticDropout rates.}
\label{fig:static_dropout_bias1}
\vspace{-0.7cm}
\end{figure}

\textbf{Results.} Figure~\ref{fig:static_dropout_bias1} shows the debiasing results for models trained on Diving48 and SSv2. StaticDropout is successful in debiasing the model away from static and toward dynamic information. In terms of our metric, (Eq.~\ref{eq:ind_bias_scores_diff_b}), both the models see a significant jump in the ratio of dynamic to static units on SSv2 and Diving48 (Fig.~\ref{fig:static_dropout_bias1} right). 
The debiasing also has a strong effect in reducing the models' ability to classify shuffled video frames compared with the standard or no-dropout baselines (Fig.~\ref{fig:static_dropout_bias1} left). 
These results suggest that StaticDropout has a strong influence on the types of specialized units contained in the model and also the model's bias toward encoding dynamic and static information. 
Figure~\ref{fig:static_dropout_bias1} (bottom) shows the results in terms of joint encoding units. Interestingly, most models contain fewer joint encoding units as the StaticDropout rate is increased, but the 3D ResNet18 trained on Diving48 deviates slightly from this pattern. This suggests that StaticDropout encourages the model to produce specialized units, in the form of static, dynamic or residual units.
Notably, the effect of StaticDropout on the model bias is consistently \textit{dose-dependant}, meaning the biases are tunable via the dropout rate, $r$.

\begin{figure} [t]
 \def\arraystretch{1.35}
 \setlength\tabcolsep{2.4pt}
\centering
\resizebox{0.4\textwidth}{!}{
	\begin{tabu}{c cccc}
\tabucline[1pt]{-}
 \multirow{2}{*}{Dropout}&  \multicolumn{2}{c}{\textbf{SSv2~\cite{goyal2017something}}} & \multicolumn{2}{c}{\textbf{SomethingElse~\cite{materzynska2020something}}}    \\
\cline{2-5}
  & Top-1 (\%) & Dyn. Units (\%) & Top-1 (\%) &  Dyn. Units (\%) \\
\tabucline[1pt]{-}
        None & 54.18 & 10.7 & 49.43  & 4.6 \\
        Normal & 54.55 & 18.2 & 49.79 &  17.5\\
        \hline
        StaticDropout  ($r$=0.1) & \textbf{55.19} & 56.9  & 50.38 &  34.2 \\
        StaticDropout ($r$=0.3) & 54.55 & 53.6  & \textbf{51.16} &  45.1 \\
        StaticDropout ($r$=0.5) & 54.01 & \textbf{80.0}  & 50.06 &  \textbf{51.4} \\
	\tabucline[1pt]{-}
	\end{tabu}}
\vfill
\resizebox{0.49\textwidth}{!}{
        \begin{tikzpicture}
        \begin{axis}[
            ybar,
            height=3.6cm,
            width=8cm,
            legend style={at={(0.5,-0.15)},
              anchor=north,legend columns=-1},
            ylabel={Ave. Acc. Diff. (\%)},
            xmin=0, xmax=55,
            ymin=-11, ymax=11,
            xtick={5,10,15,20,25,30,35,40,45,50},
            xlabel={Top-$k$ \textbf{least} common classes},
            x tick label style={font=\small},
            y tick label style={font=\small},
            x label style={at={(axis description cs:0.5,0.03)},anchor=north,font=\large},
            y label style={at={(axis description cs:0.1,.5)},anchor=south,font=\normalsize},
            nodes near coords,
            every node near coord/.append style={font=\scriptsize},
            nodes near coords align={vertical},
            extra y ticks = 0.0,
            extra y tick labels={},
            extra y tick style={grid=major,major grid style={draw=black}}
            ]
        \addplot[blue,fill=blue] coordinates {(5,1.1) (10,2.2) (15,4.1) (20,4.5) (25,4.8) (30,4.5) (35,4.6) (40,4.8) (45,4.7) (50,4.4)};

        \end{axis}
        \end{tikzpicture}
        \begin{tikzpicture}
        \begin{axis}[
            ybar,
            height=3.6cm,
            width=8cm,
            legend style={at={(0.5,-0.15)},
              anchor=north,legend columns=-1},
            xmin=0, xmax=55,
            ymin=-11, ymax=11,
            xtick={5,10,15,20,25,30,35,40,45,50},
            xlabel={Top-$k$ \textbf{most} common classes},
            x tick label style={font=\small},
            y tick label style={font=\small},
            yticklabel pos=right,
            x label style={at={(axis description cs:0.5,0.03)},anchor=north,font=\large},
            y label style={at={(axis description cs:0.1,.5)},anchor=south,font=\small},
            nodes near coords,
            every node near coord/.append style={font=\scriptsize},
            nodes near coords align={vertical},
            extra y ticks = 0.0,
            extra y tick labels={},
            extra y tick style={grid=major,major grid style={draw=black}}
            ]
        \addplot[red,fill=red] coordinates {(5,-1.1) (10,-0.7) (15,-1.2) (20,-1.3) (25,-1.0) (30,-0.9) (35,-0.5) (40,-0.3) (45,-0.2) (50,-0.3)};
        \end{axis}
        \end{tikzpicture}}

        \vspace{-0.35cm}
\caption{\textbf{Top:} StaticDropout results for 3D-ResNet50 trained on SSv2 and SomethingElse~\cite{materzynska2020something}. 
\textbf{Bottom:} Average performance difference between StaticDropout ($r=0.3$) and baseline dropout of a 3D-ResNet50 for the top-$k$ most and least common classes in the SomethingElse~\cite{materzynska2020something} compositional split. StaticDropout performs better on rare classes, while performing marginally worse on common classes. 
}\label{fig:AR_infodrop_somethingelse}
\vspace{-0.55cm}
\end{figure}
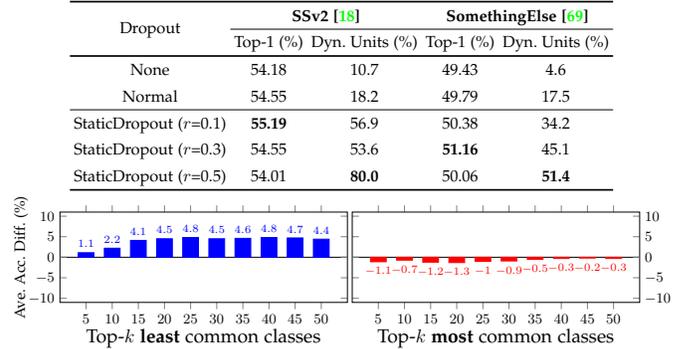

We evaluate the performance of models trained with StaticDropout on the SSv2~\cite{goyal2017something}, as it requires the maximal amount of dynamics compared with other datasets (see Sec~\ref{sec:ar_dataset}). We also evaluate on the SomethingElse~\cite{materzynska2020something} dataset, an `object debiased' relabelled version of SSv2 where the set of objects (\ie ``somethings'') appearing in the training set for a specific action is disjoint from the set of objects appearing in the validation set for the same action (see\cite{materzynska2020something} for additional details). 
We experiment on the SomethingElse compositional split with the intuition that our StaticDropout technique may debias the model from focusing on object appearances and more generally toward longer range motions contained in the videos. The results for both dataset splits are shown in Fig.~\ref{fig:AR_infodrop_somethingelse} (top) in terms of Top-1 accuracy and percentage of dynamic units in the model's final layer. StaticDropout outperforms both the baselines in terms of accuracy, and significantly increases the number of dynamic units. However, a dropout rate of $r=0.7$ does not achieve the best performance, suggesting that simply maximizing the number of dynamic units is not optimal, and that a balance of dynamics and statics should be learned.
Figure~\ref{fig:AR_infodrop_somethingelse} (bottom) shows the average performance difference between a 3D-ResNet50 trained with StaticDropout compared with standard dropout on the top-\textit{k} most and least common classes. Interestingly, on average, our model significantly outperforms the baseline on the rarest classes in the dataset while slightly under-performing the baseline on the most common classes. This result suggests that our regularization specifically targets classes in the tail end of the data distribution.

\vspace{-0.4cm}
\subsection{Fusion and Cross Connection Study}\label{sec:vos_cc_study}
\vspace{-0.1cm}
Our goal in this final section is to encourage a previously static biased AVOS model (\ie RTNet) to have more dynamic units. Based on our analysis in Sections~\ref{sec:vos_archs} and~\ref{sec:vos_datasets}, we hypothesize that the reason behind AVOS models being static biased stems from the sub-optimal application of bidirectional cross connections.
Toward this goal, we conduct a detailed analysis of the fusion and cross connection types in the aforementioned architectures and evaluate how they affect biases and model accuracy.

\begin{figure}[t]
\begin{minipage}{0.45\textwidth}
\resizebox{1.0\textwidth}{!}{
\begin{tikzpicture}
\begin{axis} [xbar stacked,
    width=\axisdefaultwidth,
    height=3.7cm,
    bar width = 8pt,
    xmin = 0,
    xmax = 100,
    title = \textbf{Fusion Layer 5},
    title style={at={(axis description cs:0.5,1.15)},anchor=north,font=\large},
    ytick=data,
    legend style={
         draw=none,
         legend style={row sep=0.1pt},
        nodes={scale=0.87, transform shape},
        legend columns=-1,
        cells={anchor=west},
        legend style={at={(0.5,1.5)},anchor=north,row sep=0.01pt}, font =\large},
    symbolic y coords={G+U, CCG+U, RTNet*/G+B, RTNet},
    enlarge x limits = {value = .1},
    enlarge y limits={abs=10pt}
]

\addplot coordinates {(37.5244140625,G+U) (0.0,RTNet) (44.2,CCG+U) (33.0,RTNet*/G+B)};
\addplot coordinates {(44.62890625,G+U) (1.171875,RTNet) (44.6,CCG+U) (47.1,RTNet*/G+B)};
\addplot coordinates {(17.8466796875,G+U) (98.828125,RTNet) (11.2,CCG+U) (20.0,RTNet*/G+B)};
\addplot coordinates {(0.0,G+U) (0.0,RTNet)(0.0,CCG+U) (0.0,RTNet*/G+B)};

\legend {Dynamic,Static,Joint,Residual};

\end{axis}
\end{tikzpicture} 
\hfill
\begin{tikzpicture}
\begin{axis} [xbar stacked,
    width=\axisdefaultwidth,
    height=3.7cm,
    bar width = 8pt,
    xmin = 0,
    xmax = 100,
    title = \textbf{Fusion Layer 4},
    title style={at={(axis description cs:0.5,1.15)},anchor=north,font=\large},
    ytick=data,
    legend style={
         draw=none,
         legend style={row sep=0.1pt},
        nodes={scale=0.87, transform shape},
        legend columns=-1,
        cells={anchor=west},
        legend style={at={(0.5,1.5)},anchor=north,row sep=0.01pt}, font =\large},
    symbolic y coords={G+U, CCG+U, RTNet*/G+B, RTNet},
    enlarge x limits = {value = .1},
    enlarge y limits={abs=10pt}
]

\addplot coordinates {(33.7890625,G+U) (0.0,RTNet)(36.6,CCG+U) (16.9,RTNet*/G+B)};
\addplot coordinates {(37.40234375,G+U) (65.6,RTNet)(37.9,CCG+U) (53.4,RTNet*/G+B)};
\addplot coordinates {(28.80859375,G+U) (34.4,RTNet)(25.5,CCG+U) (25.6,RTNet*/G+B)};
\addplot coordinates {(0.0,G+U) (0.0,RTNet)(0.0,CCG+U) (4.1,RTNet*/G+B)};

\legend {Dynamic,Static,Joint,Residual};

\end{axis}
\end{tikzpicture}
}
\vfill
\resizebox{1.0\textwidth}{!}{
\begin{tikzpicture}
\begin{axis} [xbar stacked,
    width=\axisdefaultwidth,
    height=3.7cm,
    bar width = 8pt,
    xmin = 0,
    xmax = 100,
    title = \textbf{Fusion Layer 3},
    title style={at={(axis description cs:0.5,1.15)},anchor=north,font=\large},
    xlabel = Units Encoding Factor $F$ (\%),
    xlabel style = {font=\large},
    ytick=data,
    legend style={
			area legend,
			at={(0.5,1)},
			anchor=north,
			legend columns=-1},
    symbolic y coords={G+U, CCG+U, RTNet*/G+B, RTNet},
    enlarge x limits = {value = .1},
    enlarge y limits={abs=10pt}
]
\addplot coordinates {(33.69140625,G+U) (3.9,RTNet)(37.9,CCG+U) (31.1,RTNet*/G+B)};
\addplot coordinates {(46.6796875,G+U) (64.1,RTNet)(43.9,CCG+U) (45.1,RTNet*/G+B)};
\addplot coordinates {(19.62890625,G+U) (26.6,RTNet)(18.2,CCG+U) (23.8,RTNet*/G+B)};
\addplot coordinates {(0.0,G+U) (5.5,RTNet)(0.0,CCG+U) (0.0,RTNet*/G+B)};


\end{axis}
\end{tikzpicture}
\begin{tikzpicture}
\begin{axis} [xbar stacked,
     width=\axisdefaultwidth,
    height=3.7cm,
    bar width = 8pt,
    xmin = 0,
    xmax = 100,
    title = \textbf{Fusion Layer 2},
    title style={at={(axis description cs:0.5,1.15)},anchor=north,font=\large},
    xlabel = Units Encoding Factor $F$ (\%),
    xlabel style = {font=\large},
    ytick=data,
    legend style={
			area legend,
			at={(0.5,1)},
			anchor=north,
			legend columns=-1},
    symbolic y coords={G+U, CCG+U, RTNet*/G+B, RTNet},
    enlarge x limits = {value = .1},
    enlarge y limits={abs=10pt}
]
\addplot coordinates {(29.8828125,G+U) (6.3,RTNet)(34.2,CCG+U) (30.3,RTNet*/G+B)};
\addplot coordinates {(46.2890625,G+U) (59.4,RTNet)(44.3,CCG+U) (43.8,RTNet*/G+B)};
\addplot coordinates {(23.828125,G+U) (21.9,RTNet)(21.5,CCG+U) (26.0,RTNet*/G+B)};
\addplot coordinates {(0.0,G+U) (12.5,RTNet)(0.0,CCG+U) (0.0,RTNet*/G+B)};
\end{axis}
\end{tikzpicture}}
\centering \resizebox{0.8\textwidth}{!}{
\begin{tabu}{@{}lccccc}
\tabucline[1pt]{-}
Method & Fusion & CC & Flip &mIoU & $SR_{\text{Mean}}$ \\ \tabucline[1pt]{-}
FusionSeg~\cite{jain2017fusionseg} & - & - & \xmark & 42.3 & 39.2\\
RTNet~\cite{ren2021reciprocal} & CCG & B & \xmark & 60.7 & 50.2 \\ 
MATNet~\cite{zhou2020motion} & G & U & \cmark & 64.2  & 54.4 \\\hline
MATNet* & G & U & \xmark & 67.3 & 56.6\\
MATNet* & CCG & U & \xmark& 64.4 & 51.8 \\
RTNet* & G & B & \xmark & \textbf{70.3} &  \textbf{58.1}\\\hline
MATNet* & G & U & \cmark & 68.5  & 58.1 \\
RTNet* & G & B & \cmark & \textbf{70.6} & \textbf{58.6}\\ \hline
\tabucline[1pt]{-}
\end{tabu}}
\end{minipage}
\vspace{-0.2cm}
    \caption{\textbf{Top:} Estimates of the dynamic, static, joint and residual units using our metric for unit-wise analysis, (3), $\lambda=0.5$ at different fusion layers of RTNet. \textbf{Bottom:} Evaluation of AVOS models on MoCA~\cite{lamdouar2020betrayed}. 
    * means networks implemented by us (see Sec.~\ref{sec:vos_archs}).
    We compare cross connections: unidirectional motion-to-appearance (U) \vs bidirectional (B), and fusion types: gated (G) \vs convex combination gated (CCG). Flip means performing data augmentation with flipping during inference, as originally done in~\cite{zhou2020motion}.}
    \label{fig:moca_vs_davis_and_ind}
    \vspace{-0.6cm}
\end{figure}
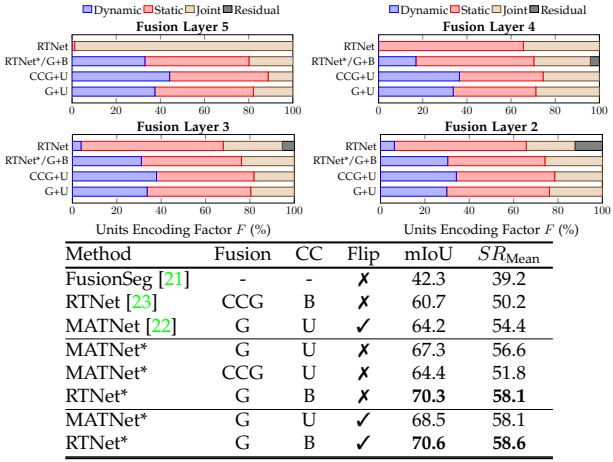

\noindent \textbf{Approach.} We now analyze the different types of \textit{fusion} and \textit{cross connections}. Cross connections can operate \textit{Bidirectionally} (\ie motion-to-appearance and appearance-to-motion), as in RTNet, or \textit{Uni-directionally} with motion-to-appearance cross connection, as in MATNet. The \textit{Convex Combination Gated Fusion} in RTNet enforces the attention weights to sum to one for both motion and appearance streams. While the \textit{Gated Fusion} in MATNet learns to weigh each stream without this constraint. Both mechanisms can be decomposed into channel attention followed by spatial attention. For the sake of unification we pose the channel attention in the \textit{Convex Combination Gated Fusion} as,
\begin{subequations}
\begin{gather}
    A_c = \mathcal{F}_e(\mathcal{F}_s(U_a \oplus U_m)),  \quad   Z_a = A_c \circ U_a,  \tag{\theequation a,b} \\
 Z_m = (1-A_c) \circ U_m, \tag{\theequation c}
\end{gather}
\end{subequations}
where $U_a, U_m$ are the appearance and motion features, respectively, $\circ$ is the Hadamard product, $\oplus$ denotes concatenation, $\mathcal{F}_s$ and $\mathcal{F}_e$ are the squeeze and excitation operators, respectively, and $A_c \in \mathbb{R}^c$ are the channel attention weights where $c$ is the number of channels for the input appearance or motion feature maps. A weighted combination is used to attend to motion and appearance features, enforcing a strong constraint on the attention weights during training. On the other hand, the channel attention module in the \textit{Gated Fusion} can be given as
\vspace{-0.1cm}
\begin{subequations}
\begin{gather}
U = U_a \oplus U_m,  \quad  A_c = \mathcal{F}_e(\mathcal{F}_s(U)),   \quad   Z = A_c \circ U, \tag{\theequation a-c} 
\end{gather}
\end{subequations}
\vspace{-0.1cm}
where $A_c \in \mathbb{R}^{2c}$ is the channel-wise attention on the combined motion and appearance features without enforcing them to to sum to one, unlike in RTNet.

In contrast, the spatial attention in \textit{Convex Combination Gated Fusion} can be given as
\vspace{-0.1cm}
\begin{equation} 
Z = A_{sp} \circ Z_a \oplus (1-A_{sp}) \circ Z_m,
\end{equation}
\vspace{-0.1cm}
where $A_{sp}$ is the spatial attention maps generated from a convolutional layer and a sigmoid function, while $Z_a$ and $Z_m$ are the average pooled appearance and motion features. Finally, the spatial attention in \textit{Gated Fusion} can be given as
\vspace{-0.1cm}
\begin{equation}
 Z = A_{sp} \circ Z + Z,
\end{equation}
\vspace{-0.1cm}
\noindent where $Z$ is the combined motion and appearance features. Again, it is clear that the spatial attention module in the \textit{Convex Combination Gated Fusion} restricts the attention weights between both motion and appearance features, unlike the \textit{Gated Fusion}. Therefore we hypothesis that \textit{Gated Fusion} will yield better results for tasks requiring a complex interaction between motion and appearance features. We now ablate the fusion types along with the two different types of cross connections on such a task.

\noindent \textbf{Evaluation protocol.} We show the static and dynamic biases for the different fusion and cross connections, and evaluate accuracy on a dataset which requires dynamics (MoCA). We follow previous work by removing videos that contain no predominant target locomotion, which produces a subset of 88 videos for evaluation~\cite{yang2021selfsupervised}. We evaluate using mean intersection over union and success rate mean with varying IoU thresholds ranging from 0.5 to 0.9. The original MATNet used horizontal flipping during the inference and averaged predictions from the original and flipped versions. To ensure fair comparison between RTNet and FusionSeg we show results with and without the flipping augmentation during inference when reporting on MoCA.

\noindent\textbf{Results.}
Figure~\ref{fig:moca_vs_davis_and_ind} (top) shows the static and dynamic biases of the different fusion and cross connection types. It is seen that the bidirectional cross connections (G+B) incur a small decrease in the dynamic bias with respect to unidirectional ones (G+U), especially on the final fusion layers (\ie fusion 4 and 5). The convex combination gated fusion (CCG+U) leads to a decrease in the ratio of joint to dynamic units with respect to gated fusion (G+U) (\eg the final fusion layer for CCG+U Joint/Dynamic is 25.3 \vs 47.5 for G+U). The performance of all models on the MoCA dataset is shown in Fig.~\ref{fig:moca_vs_davis_and_ind} (bottom). The original MATNet and MATNet* both outperform the off-the-shelf RTNet model and FusionSeg, which is confirmation of our previous findings that both models are heavily static biased (see Fig.~\ref{fig:stagewise_vos}). Additionally, it demonstrates that our proposed model (RTNet*) with bidirectional cross connections and Gated Fusion (B+G) trained on YouTube-VOS and DAVIS shows sizable ($+9.9\%$) gain with respect to off-the-shelf RTNet. We also observe from the results of MATNet* (G+U) \vs MATNet* (CCG+U) that the added constraint of convex combination can degrade the performance with respect to gated fusion on a dynamically heavy task (\ie MoCA).

Motivated by the previous experiments. which showed the impact of static and dynamic biases on performance (Sec.~\ref{sec:neuron_removal}), these results demonstrate how simply selecting the appropriate combination of fusion and cross connection mechanisms can significantly impact what is encoded in models as well as the downstream performance.

\section{Conclusion}

This paper has advanced the interpretability of learned spatiotemporal models for video understanding, especially action recognition, AVOS and VIS. We introduced a general method for analyzing the extent that various architectures capitalize on static \vs dynamic information. We showed how our method can be applied to investigate the static \vs dynamic biases in datasets. Furthermore, we demonstrated the impact of static and dynamic biases on overall performance through a new type of regularization for action recognition (StaticDropout) and architectural modifications in the fusion and cross connection layers for AVOS. Future work can apply our method to additional video understanding tasks (\eg action prediction) and use the proposed performance enhancing techniques on different datasets, architectures and tasks which require dynamics.

\bibliographystyle{IEEEtran}
\bibliography{main}

%
\begin{IEEEbiography}[{\includegraphics[width=1.1in,height=1.5in,clip,keepaspectratio]{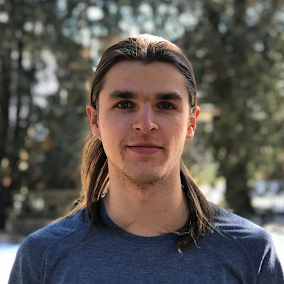}}]{Matthew Kowal}
received the Bachelor of Applied Science degree in Applied Math from Queen's University, Canada in 2017. He received the MSc degree in computer science from Ryerson University, Canada in 2020. He is currently a PhD student in the department of Electrical Engineering and Computer Science in York University, Toronto, Canada and is a Technical Lead at the Vector Institute. Previously, he held the position of Lead Scientist in Residence at NextAI and interned at Toyota Research Institute and Ubisoft La Forge. His main research field of interest is in the intersection of interpretable computer vision and video understanding.
\end{IEEEbiography}
\vspace{-20pt}
\begin{IEEEbiography}[{\includegraphics[width=1.1in,height=1.5in,clip,keepaspectratio]{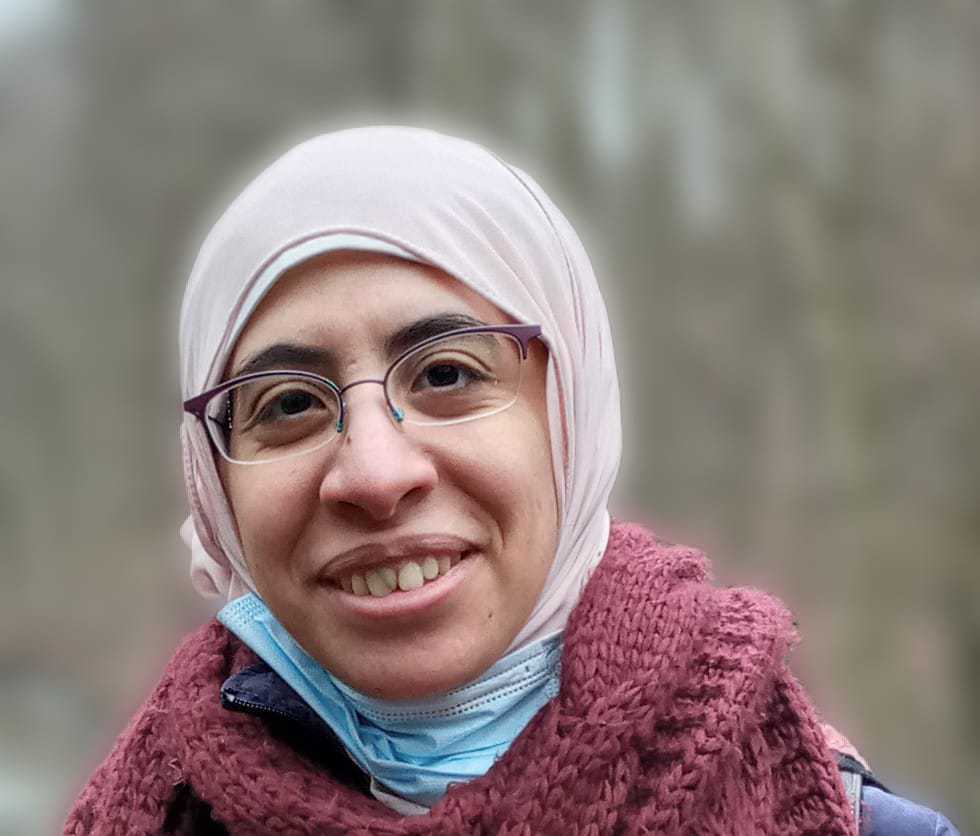}}]{Mennatullah Siam} is an assistant professor in Ontario Tech University since 2023 and an affiliate professor in University of British Columbia. She is leading the Image and Video Understanding (IVU) lab. Previously, She was a postdoctoral researcher in the department of Electrical Engineering and Computer Science in York University, Toronto, Canada and a Vector institute affiliate. She received the B. Sc. degree from Computer Science in Ainshams University, Cairo in 2010, the MSc degree in Informatics from Nile University, Egypt in 2013 and the PhD degree from Computing Science department in University of Alberta, Edmonton, Canada in 2021. She is a recipient of multiple NSERC grants and Canada Access Alliance, and she was a recipient of Alberta Innovates Foundations Technology scholarship, the Verna Tate graduate scholarship, and the VISTA postdoc fellowship. Her major fields of interest are Computer Vision, Deep Learning and Responsible AI, where she is focusing on video understanding, interpretability, data efficient learning and their societal impact on marginalized communities.
\end{IEEEbiography}
\vspace{-20pt}
\begin{IEEEbiography}[{\includegraphics[width=1.1in,height=1.5in,clip,keepaspectratio]{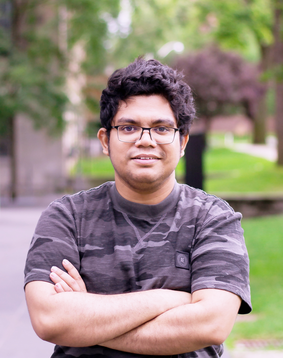}}]{Md Amirul Islam} is a Senior Researcher at Noah's Ark Lab, Huawei Technologies Canada. He received his PhD in computer science from Ryerson University, Canada in 2022. He was a Postgraduate Affiliate at the Vector Institute. He received his M.Sc. in Computer Science from University of Manitoba, Canada in 2017 and his B.Sc. in Computer Science and Engineering from
North South University, Bangladesh in 2014. His research interests are in human-centric AI and multi-modal computer vision. 
\end{IEEEbiography}
\vspace{-20pt}
\begin{IEEEbiography}[{\includegraphics[width=1.1in,height=1.5in,clip,keepaspectratio]{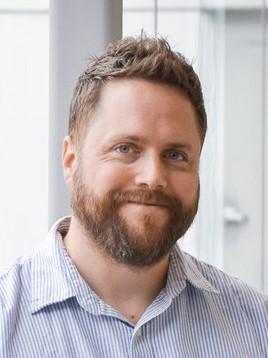}}]{Neil D. B. Bruce} graduated from the University of Guelph with a B.Sc. Double major in CS and Pure Mathematics. Dr. Bruce then attended the University of Waterloo for an M.A.Sc. in System Design Engineering and York University for a Ph.D. in Computer Science. Prior to joining Guelph he worked in the Department of Computer Science at Ryerson University. Prior to this Dr. Bruce worked at the University of Manitoba as Assistant then Associate Professor. Dr. Bruce has postdoctoral experience working at INRIA (France) and Epson Canada. He is the recipient of the Falconer Rh Young Researcher Award and is a Faculty Affiliate at the Vector Institute, Toronto. His research explores solutions to issues in computer vision, deep-learning, human perception, neuroscience and visual computing.
\end{IEEEbiography}
\vspace{-20pt}



\begin{IEEEbiography}[{\includegraphics[width=1in,height=1.25in,clip,keepaspectratio]{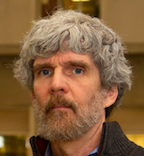}}]{Richard P. Wildes}
Richard P. Wildes (Member, IEEE) received the PhD degree from the Massachusetts Institute of Technology in 1989. Subsequently, he joined Sarnoff Corporation in Princeton, New Jersey, as a Member of the Technical Staff in the Vision Technologies Lab. In 2001, he joined the Department
of Electrical Engineering and Computer
Science at York University, Toronto, where he is
a Professor, a member of the Centre for Vision
Research and a Tier I York Research Chair. He also is a visiting research scientist at Samsung Artificial
Intelligence Center (SAIC), Toronto.
Honours include receiving a Sarnoff Corporation
Technical Achievement Award, the IEEE D.G. Fink Prize Paper Award
and twice giving invited presentations to the US National Academy of
Sciences. His main areas of research interest are computational vision,
especially video understanding, and artificial intelligence.
\end{IEEEbiography}
\vspace{-20pt}

\begin{IEEEbiography}[{\includegraphics[width=1in,height=1.25in,clip,keepaspectratio]{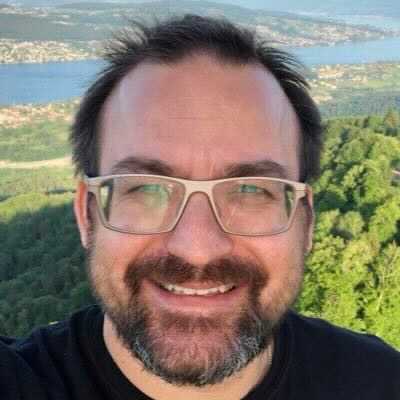}}]{Konstantinos G. Derpanis}
Kosta Derpanis received the Honours Bachelor of Science (BSc) degree in computer science with a Minor in mathematics from the University of Toronto, Canada, in 2000.  He received the MSc (supervisors Prof. John Tsotsos and Prof. Richard Wildes) and PhD (supervisor Prof. Richard Wildes) degrees in computer science from York University, Canada, in 2003 and 2010, respectively.  For his dissertation work, he received the Canadian Image Processing and Pattern Recognition Society (CIPPRS) Doctoral Dissertation Award 2010 Honourable Mention.  Subsequently, he was a postdoctoral researcher in the GRASP Laboratory at the University of Pennsylvania under the supervision of Prof. Kostas Daniilidis.  In 2012, he joined the Department of Computer Science at Ryerson University, Toronto, and later was promoted to Associate Professor with early tenure. In 2021, he joined the Department of Electrical Engineering and Computer Science at York University as an Associate Professor.  He also serves as a visiting research scientist at Samsung Artificial Intelligence Center (SAIC), Toronto. His main research field of interest is computer vision with emphasis on motion analysis and human motion understanding, and related aspects in image processing and machine learning.
\end{IEEEbiography} 




\end{document}